\documentclass[%
 reprint,
groupedaddress,
 bibnotes,
 amsmath,
 amssymb,
 aps,
 prl,
]{revtex4-2}

\usepackage{graphicx}

\newcommand{\methodName}[0]{Neptune}
\usepackage{color,soul, colortbl}
\usepackage{mathtools}
\usepackage{bm}
\usepackage{amsmath}

\usepackage{ulem}
\usepackage{graphicx}
\usepackage{orcidlink}
\usepackage{adjustbox}
\usepackage{multirow}
\usepackage{booktabs}
\usepackage{mathrsfs}
\usepackage{threeparttable}
\usepackage{hyperref}
\usepackage{pdfpages}

\makeatletter
\AtBeginDocument{\let\LS@rot\@undefined}
\makeatother

\usepackage{xcolor}
\hypersetup{
    colorlinks,
    linkcolor={blue},
    citecolor={blue},
    urlcolor={blue}
}

\usepackage{etoolbox}
\patchcmd{\section}
  {\centering}
  {\raggedright}
  {}
  {}

\begin{document}
\title{Estimating Parameter Fields in Multi-Physics PDEs from Scarce Measurements}
\author{Xuyang Li\orcidlink{0000-0002-6846-0906}}
\author{Mahdi Masmoudi\orcidlink{0009-0003-3326-9305}}
\author{Rami Gharbi\orcidlink{0009-0002-6012-1122}}
\author{Nizar Lajnef\orcidlink{0000-0001-9578-1054}}
\author{Vishnu Naresh Boddeti\orcidlink{0000-0002-8918-9385}}
\affiliation{Michigan State University, East Lansing, MI 48824, USA}
\date{\today}

\begin{abstract}
Parameterized partial differential equations (PDEs) underpin the mathematical modeling of complex systems in diverse domains, including engineering, healthcare, and physics. A central challenge in using PDEs for real-world applications is to accurately infer the parameters, particularly when the parameters exhibit non-linear and spatiotemporal variations. Existing parameter estimation methods, such as sparse identification, physics-informed neural networks (PINNs), and neural operators, struggle in such cases, especially with nonlinear dynamics, multiphysics interactions, or limited observations of the system response. To address this, we introduce \methodName{}, a {\color{black}versatile} method capable of inferring parameter fields from sparse measurements of system responses. \methodName\ employs independent coordinate neural networks to continuously represent each parameter field in physical space or in state variables. Across various physical and biomedical problems, where direct parameter measurements are prohibitively expensive or unattainable, \methodName\ significantly outperforms existing methods, achieving robust parameter estimation from as few as 45 measurements, reducing parameter estimation errors by up to two orders of magnitude and dynamic response prediction errors by a factor of ten to baseline methods. More importantly, it exhibits superior physical extrapolation capabilities, enabling reliable predictions in regimes far beyond the training data. By facilitating data-efficient parameter inference, \methodName\ promises {\color{black}significant utility} in engineering, healthcare, and beyond.
\end{abstract}
\maketitle

\begin{figure*}[!ht]
\centering
\includegraphics[width=\textwidth]{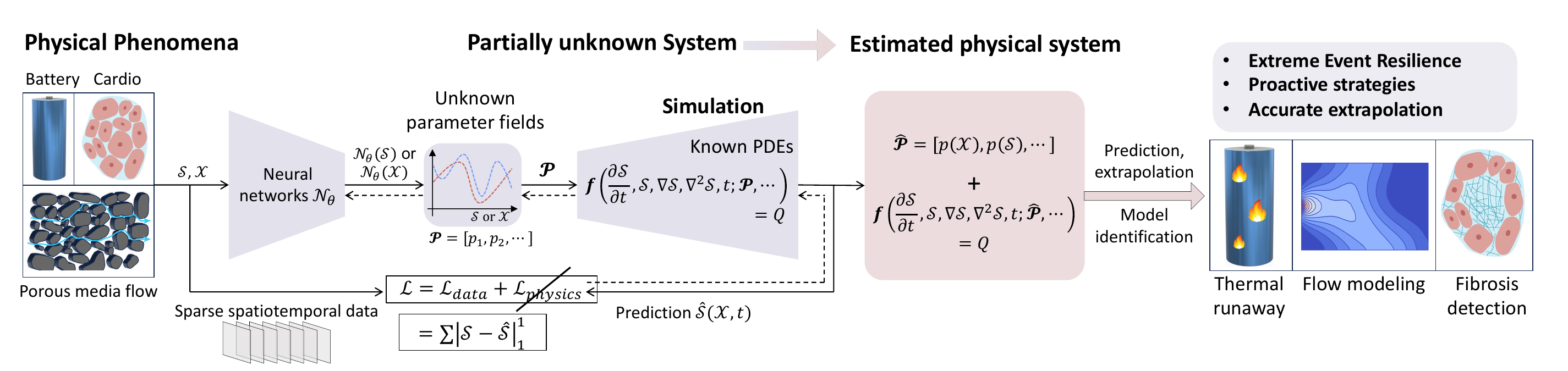}
\caption{\textbf{Overview of \methodName}, a {\color{black}versatile} method for inferring unknown parameter fields from sparse observations of system responses. Physical phenomena are often governed by known PDEs or coupled PDE systems with unknown key parameter fields. \methodName\ estimates these complex fields using extremely sparse physical response data through a two-stage process to ensure robustness. First, scalar parameters are estimated under an assumed homogeneous field approximation. Second, neural networks (referred to as $\mathcal{N}$) model the local variance of parameter fields, along with the previous scalar estimations. {\color{black}Here, $S$ denotes the physical state variables (e.g., temperature or concentration), $P$ represents the unknown parameter fields to be estimated, and $X$ indicates the spatial coordinates.} Both stages iteratively solve the PDE system numerically, minimizing the error between predicted and observed dynamics via adjoint-based backpropagation. The trained method yields a reliable physical system model capable of predicting phenomena under varying environmental conditions (e.g., source terms, boundary conditions), enabling Extreme Event Resilience, Accurate Extrapolation, and Proactive Strategies as demonstrated in predicting battery thermal runaway behavior, characterizing flow in porous media, and detecting cardiac fibrosis via electrophysiological analysis.\label{fig:1} }
\end{figure*}

Accurate modeling of physical phenomena is crucial in a wide range of scientific and engineering applications. Differential equations provide a foundational framework for capturing dynamics that evolve over time and space. 
However, the governing parameters are often unknown or deviate from known values due to factors such as aging, wear, or environmental variation, resulting in significant changes in system behavior.
For example, the thermal properties of aged batteries can evolve, increasing the risk of thermal runaway and fires in electric vehicles (EVs). 
Similarly, in biomedical contexts, changes in cardiac tissue properties can lead to arrhythmias or atrial fibrillation, while variations in porous media flow properties may cause groundwater contamination or aquifer depletion.
These examples underscore the critical need for parameter estimation, as failing to account for evolving parameters can lead to inaccurate predictions and severe consequences. 
Beyond accuracy, parameter estimation enables {\color{black}many downstream} applications, such as simulating future behavior, predicting failure points, estimating remaining useful life in engineering systems, and creating personalized models for tailored healthcare diagnostics and treatments through high-fidelity digital twins.

Parameter estimation in multiphysics applications presents several interconnected challenges. The problem is ill-posed due to its inverse nature. In many real-world scenarios, direct parameter measurement is impractical due to cost, inaccessibility, or physical constraints, leaving system responses as the only observable data. In addition, available system response data are often sparse, noisy, or incomplete, limiting reliable parameter recovery. More critically, many relevant parameters are not scalar but spatially distributed or field-dependent, and are often tightly coupled across scales, making them difficult to isolate or model independently.

Despite notable progress in parameter estimation methods, significant limitations remain, particularly when addressing nonlinear multiphysics systems or time-dependent behaviors. Traditional approaches such as finite element updating~\cite{steenackers2006finite, ebrahimian2017nonlinear}, Bayesian neural networks~\cite{yang2021b}, least squares estimation~\cite{ji2020hierarchical, li2020maximum}, Kalman filtering~\cite{varshney2019state, hossain2022kalman}, Gaussian processes~\cite{zhang2022parameter, deng2020data, li2020state}, and sparse identification~\cite{brunton2016discovering, chen2021physics} are effective in simplified, single-physics problems. However, they struggle with spatially and temporally varying parameters in nonlinear multiphysics systems. Karhunen–Loève expansions~\cite{tipireddy2020conditional, tartakovsky2021physics, li2022physics} can represent parameter fields, but have limited applicability to time-dependent processes~\cite{tipireddy2020conditional, tartakovsky2021physics, he2020physics} and often require prior assumptions about parameter distributions~\cite{li2022physics}.

Recent machine learning based approaches, such as PINNs~\cite{toscano2025pinns, karniadakis2021physics, raissi2019physics, hansen2023learning, wang2024respecting, pang2019fpinns, wang2025gradient, subramanian2023adaptive, krishnapriyan2021characterizing, mishra2023estimates, hu2021extended, nghiem2023physics, qian2020lift, zubov2021neuralpde, cuomo2022scientific}, Neural Operators~\cite{lu2021learning, li2020fourier, wang2022improved, jiao2021one, zhang2024blending, li2020neural, wang2021learning, kovachki2023neural, guibas2021adaptive}, and other deep surrogates~\cite{pestourie2023physics, wang2024bayesian, franco2023deep} have emerged as efficient alternatives for modeling complex, field-dependent parametric systems. 
However, growing evidence~\cite{mcgreivy2024weak} shows that these methods often rely on weak numerical baselines and suffer from reporting biases, leading to overoptimistic claims of performance and generalization.
PINNs embed governing equations into neural network loss functions, enabling simultaneous learning of system dynamics and parameter fields~\cite{gao2022physics, yuan2022pinn, jagtap2020conservative, lu2021physics, raissi2019physics, yu2022gradient}. While effective in simple cases, PINNs struggle to train or recover complex parameters from extremely sparse or noisy data, particularly in nonlinear, coupled multiphysics systems, due to the dual objectives of simultaneous forward and inverse modeling~\cite{herrero2022ep, he2020physics, he2022physics, tartakovsky2020physics, taneja2022feature}. PINN-SR~\cite{chen2021physics}, which incorporates sparse regression principles (SINDy)~\cite{brunton2016discovering} for parameter and equation discovery, faces challenges in generalizing to systems with parameter fields.
Neural Operators, such as DeepONet~\cite{lu2021learning} and Fourier Neural Operators (FNO)~\cite{li2020fourier}, learn solution maps directly from data and have been extended to inverse problems~\cite{long2024invertible, zhou2024parameter, molinaro2023neural, behroozi2025sensitivity}, yet often require large, high-fidelity datasets comprising thousands of samples~\cite{azizzadenesheli2024neural, li2024physics} and are sensitive to training data distribution~\cite{lin2023learning} and hyperparameter choices~\cite{kovachki2023neural, lin2023learning, azizzadenesheli2024neural}.

A separate line of work builds on the differential equation framework by embedding neural networks into the governing dynamics, as in neural ODEs~\cite{chen2018neural, rackauckas2019diffeqflux, dandekar2020bayesian, chen2021eventfn, kidger2022neural, lee2021parameterized}, universal differential equations~\cite{rackauckas2020universal}, {\color{black}and differentiable PDE-constrained inverse modeling frameworks spanning computer graphics, solid mechanics, and subsurface flow~\cite{xu2020adcme, liang2019differentiable, du2026jax, lin2025differentiable, xue2023jax}.}
Coupled with adjoint methods~\cite{rackauckas2020universal, rackauckas2019diffeqflux}, they enable memory‑efficient parameter estimation. 
{\color{black}Yet these differentiable PDE-constrained frameworks have primarily been demonstrated with moderately dense observations or domain-specific priors, and the neural ODE literature largely targets low-dimensional systems with scalar parameters}~\cite{yang2025neural,kong2022dynamic,bradley2021two,norcliffe2021neural}. One underlying reason is the high dimensionality and ill‑posedness of field estimation, where directly optimizing all local parameters leads to an enormous search space and unstable convergence. 
These challenges highlight the need for structured estimation strategies that can first capture dominant system behavior, before resolving more localized variations.

\begin{table*}[!ht]
\centering
\resizebox{\textwidth}{!}
{
\begin{tabular}{llllllllccc}
    \toprule
     \multirow{2}{*}{\textbf{Application}} && \multirow{2}{*}{\textbf{Physics}} && \multirow{2}{*}{\textbf{Parameters}} && \multirow{2}{*}{\textbf{Measurements}} && \textbf{Parameter} && \textbf{Extrapolated}\\
     && && && && \textbf{Error} && \textbf{Response Error} \\
     \midrule
     Battery && Thermal \& chemical && $k$ (conduct.), $C_p$ (heat cap.) && $T$ (temperature) && 0.81\%, 1.23\% && 0.36\% \\
     Porous Flow && Darcy \& dispersion && $K$ (hydraulic conduct.) && $u$ (concentration) && 8.83\% && 0.69\% \\
     Cardiac Electro && Aliev-Panfilov && $D$ (tissue conduct.) && $V$ (voltage) && 3.01\% && 0.46\% \\
     Cell Migration && Diffusion-reaction && $\gamma$ (diffusion), $\lambda$ (growth) && $\rho$ (density) && - && 6.40\% \\
    \bottomrule
\end{tabular}
}
\caption{Parameter estimation performance for a range of problems. Parameter error is defined as normalized mean absolute error{\color{black}, which allows for comparisons across applications with different physical units}, and extrapolated response error uses peak percentage error. Metrics are derived from approximately 125 measurements. Battery conductivity errors are reported for the $x$-$y$ direction only (similar in $z$). Cell migration (152 measurements, no parameter references) excluded parameter error, and the response error is obtained from the training region. {\color{black}Absolute errors for individual fields under varying noise levels are reported in Table~\ref{tab:table2}.} }
\label{tab:table1}
\end{table*}

\begin{table*}[!ht]
    \centering
    \begin{adjustbox}{max width=0.98\textwidth}
    \begin{tabular}{lllllllll}
         \toprule
         \textbf{Problem} && \textbf{Err. (N-0\%)} && \textbf{Err. (N-1\%)} && \textbf{Err. (N-10\%)} && \textbf{Description} \\
         \midrule
          Battery && 0.01 ($\pm$0.006) && 0.05 ($\pm$0.009) && 0.15 ($\pm$0.038) && $k(T)$ \\
          Porous Media Flow && 0.089 ($\pm$0.019) && 0.10 ($\pm$0.016) && 0.21 ($\pm$0.073) && $K(x,y)$\\
          Cardiac Electrophysiology && 0.0030 ($\pm$0.00082) && 0.016 ($\pm$0.0060) && 
          0.024 ($\pm$0.0052) && $D(x,y)$\\
          \bottomrule
    \end{tabular}
    \end{adjustbox}
    \caption{Summary of the \methodName\ results in the context of accuracy for a range of models. The error is defined as the average absolute error of the estimated parameter compared to the reference. 
    {\color{black}As in Table~\ref{tab:table1}, all cases use a common budget of around 125 measurements.} The values in the parentheses denote one standard deviation across 10 independent runs.
    In the battery problem, only the thermal conductivity $k$ estimation is displayed for simplicity. Absolute error is used instead of percentage error, as values near zero can distort the error assessment.}
    \label{tab:table2}
\end{table*}

To address these limitations, this article introduces \methodName\ (Neural Estimation of Parameters in mulTi-physics PDEs Under sparse obSErvations), a {\color{black}versatile} and robust method for parameter field estimation in complex systems (see Figure \ref{fig:1}). 
{\color{black}Building on our preliminary work~\cite{li2024estimating},} \methodName\ extends the universal differential equations framework from simple ODEs to multiphysics and coupled PDEs using finite difference discretization, while remaining compatible with standard numerical solvers. \methodName\ still utilizes adjoints for optimization, but it combines known governing equations with coordinate neural networks to infer unknown parameter fields, and requires no assumptions about their prior distributions.
A core feature of \methodName\ is a structured two-stage estimation strategy: first recovering coarse scalar parameters, then refining spatially-varying fields via neural networks. This enables scalable recovery of multiple, high-resolution parameter fields in coupled systems.

\methodName\ inherits the flexibility of standard numerical solvers by discretizing governing PDEs into ODE systems via finite differences, enabling direct integration into existing simulation workflows.
It does not rely on any prior assumptions about parameter distributions, and remains robust across a wide range of nonlinear, coupled multiphysics systems.
By solving directly from the governing equations, \methodName\ demonstrates strong generalization and extrapolation capabilities, delivering accurate predictions beyond the training region and across diverse physical domains.
These properties allow \methodName\ to achieve accurate and data-efficient parameter estimation, even under extremely sparse and noisy observations. Our key contributions are as follows:

\begin{itemize}
\item \textbf{{\color{black}Versatile} framework for multiphysics PDEs.} 
    \methodName\ extends universal differential equations to high-dimensional, coupled systems, without requiring prior assumptions on parameter distributions.

    \item \textbf{Two-stage parameter field estimation.} 
    A coarse-to-fine strategy enables scalable recovery of complex, spatially-varying parameter fields.

    \item \textbf{Strong generalization from physics priors.} 
    Leveraging known governing equations, \methodName\ extrapolates reliably beyond training regions and under sparse observations.
\end{itemize}

In the following sections, we demonstrate \methodName's versatility and efficacy across various scientific domains, including cardiac modeling, thermal-chemical battery systems, porous media flow, and cell migration. These applications showcase \methodName's robustness under scarce and noisy data, adaptability to complex real-world settings, and strong predictive accuracy. Quantitative results in Tables~\ref{tab:table1} and~\ref{tab:table2} highlight \methodName’s consistent performance in parameter estimation across tasks.
{\color{black}Each case study is benchmarked against baselines: PINN (and its sparse-regression variant PINN-SR) serves as the primary physics-informed baseline throughout; and PINN-SR is used for cell migration to ensure consistent comparison with the reference study analyzing this experimental dataset; the supervised operator-learning (FNO) setup requires paired training samples spanning a well-defined distribution of parameter fields, which is most tractable for the cardiac case; spline-based parameter fitting is applied to the flow problem as a classical non-neural baseline for spatial parameter mapping. A consolidated comparison, together with computational cost and an ablation of the two-stage strategy, is presented after the case studies.}

\begin{figure*}[!ht]
\centering
\includegraphics[width=0.975\textwidth]{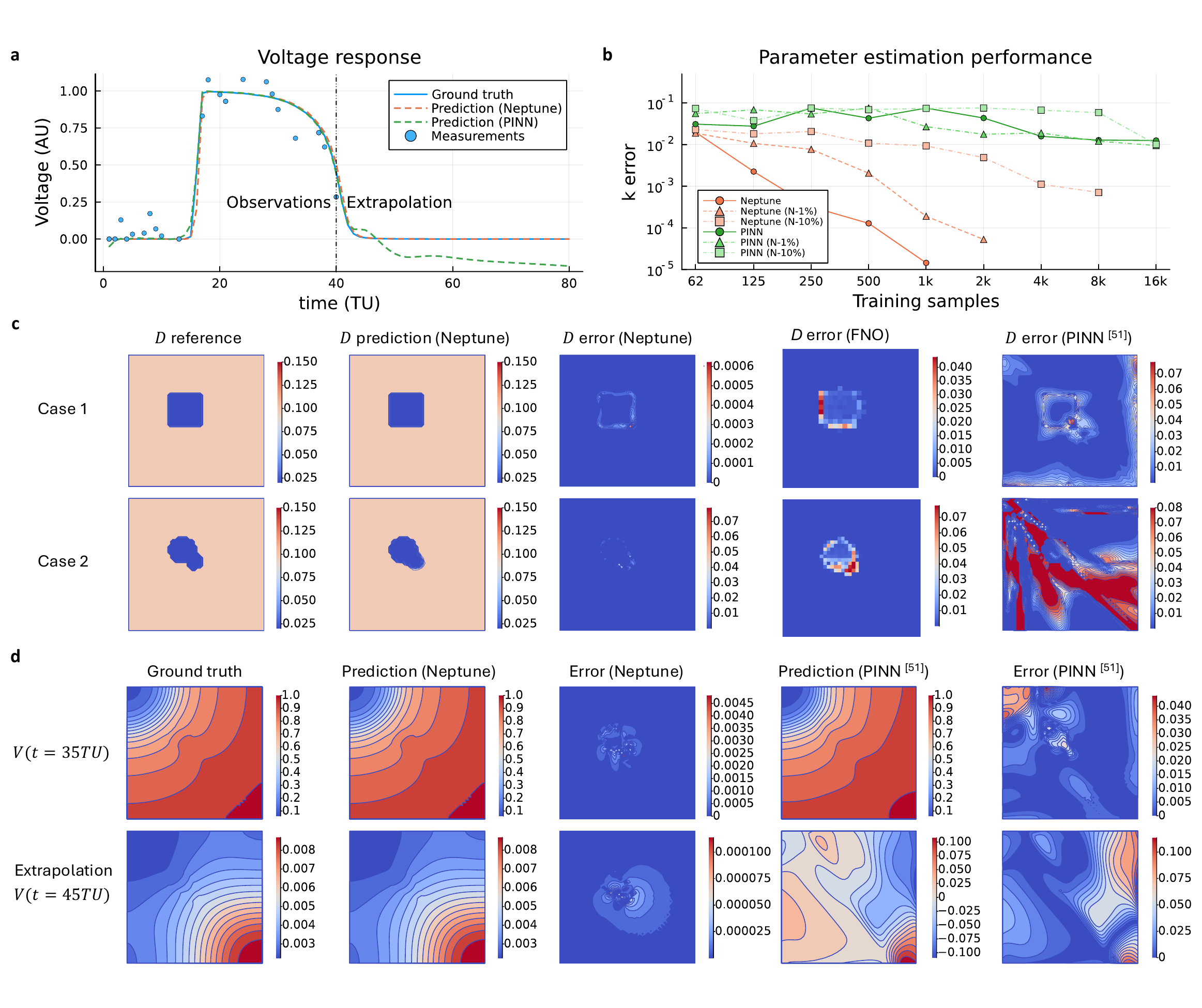}
\caption{\textbf{Parameter estimation in cardiac electrophysiology.} \textbf{a}. The training region with Gaussian noise during measurement, and the estimated forward inference of $V$. The predictions are compared between \methodName\ and PINN. \textbf{b}. The parameter estimation error under varying training data sizes. \textbf{c}. Comparison of parameter estimation error plots among \methodName, FNO, and PINN for two representative cases of $D$.  \textbf{d}. The estimated forward response $V$ comparisons for case 1. When performing temporal extrapolation, PINN predicts values that do not adhere to physics laws.\label{fig:2} }
\end{figure*}

\section{Cardiac Electrophysiology}
Cardiac electrophysiology (EP) ~\cite{kashtanova2023simultaneous, kashtanova2022aphyn, werneck2023replacing, herrero2022ep} emphasizes the crucial coupling between cardiac tissue properties and the generation as well as propagation of electrical signals within the heart.

\noindent\textbf{Known Physics:} The canine ventricular Aliev-Panfilov model is utilized in this section, with the coupled equations~\cite{herrero2022ep} defined in the following.
\begin{align} \label{eq:heart}
\frac{\partial V}{\partial t} = \nabla ( D \nabla V) - k_0V(V-a)(V-1)-VW \\
\frac{\partial W}{\partial t} = \big(\epsilon + \frac{\mu_1W}{V+\mu_2}\big)\big(-W-k_0V(V-b-1)\big)
\end{align}

This model is commonly used to identify tissue heterogeneities, particularly from in silico data~\cite{zaman2021fast, ntagiantas2024estimation, herrero2022ep}, with significant clinical potential for detecting fibrosis and other localized pathologies linked to arrhythmias and atrial fibrillation~\cite{herrero2022ep, ntagiantas2024estimation}. The diffusion tensor $D$, the target parameter, determines propagation speed, with $V$ as the measurable voltage and $W$ as an intermediate state variable. The PDEs describe the evolution of electrical potential and ionic currents in cardiac tissue, incorporating known properties such as tissue conductivity and reaction-diffusion dynamics.

\noindent\textbf{Unknown Parameters:} The key parameter, tissue conductivity (proportional to the diffusion tensor $D$), varies spatially and is challenging to measure directly, requiring estimation for accurate modeling. For simplicity, the estimation focuses on 
$D$. Details of all known physical parameters and conditions relevant to this problem, along with those for other cases discussed in the article, can be found in \textit{Supplementary Note 1}.

\noindent\textbf{Problem Setup:} A 2D slab of cardiac tissue is employed ($1cm\times1cm$) as the spatial domain and aims to recover the heterogeneous diffusion parameter $D(x,y)$. Healthy tissue is represented by $D = 0.1 mm^2/TU$, while fibrosis tissue has $D = 0.02 mm^2/TU$ five times smaller. $1TU$ is approximately $13ms$ \citep{herrero2022ep}. During training, Gaussian noise is added to the training data ($V$ measurements) to mimic real-world situations. The voltage response over time (from 0 to 40 $TU$) at a randomly selected spatial point, including noise, is shown in Fig. \ref{fig:2}a.

\noindent\textbf{Measurements (with details and comparison to baselines):}
\methodName\ demonstrates exceptional efficiency and precision in joint parameter estimation and dynamic prediction. It achieves highly accurate and stable parameter estimates with as few as 1,000 measurements, maintaining robust performance even when utilizing only 62 sparse measurements on a $50 \times 50$ mesh. Furthermore, \methodName\ significantly outperforms alternative approaches under noisy conditions, demonstrating superior accuracy and stability under challenging, data-limited scenarios. In contrast, the PINN model~\cite{herrero2022ep} requires a substantially higher data density to achieve functional results, needing at least 8,000 spatiotemporal measurements on a much finer $100 \times 100$ mesh {\color{black}before the estimated field becomes visually distinguishable}. As illustrated in Figure \ref{fig:2}b, PINN consistently exhibits higher estimation errors across all data scales. Beyond estimation accuracy, \methodName\ maintains physical consistency during temporal extrapolation, whereas PINN tends to generate predictions that violate underlying physiological laws (Figure \ref{fig:2}d). {\color{black}
Finally, the FNO model was evaluated as a direct data-driven inverse operator. Specifically, spatiotemporal measurements of the state variable $V$ on a $30 \times 30$ grid across 3 time steps (totaling 2,700 measurements) were used as the direct input to the FNO to output the spatial distribution of the parameter $D$. The model is trained in a supervised manner on paired samples of diverse geometric distributions of $D$ and their corresponding $V$ responses.}
While FNO provides a standard operator-learning benchmark, \methodName\ demonstrates a superior ability to recover parameters $D$ from significantly sparser and noisier measurement sets, as evidenced by the error plots in Figure \ref{fig:2}c.

\noindent\textbf{Analysis:} Figure \ref{fig:2}c evaluates the parameter estimation performance of \methodName, PINN, and FNO across two cases with distinct parameter distributions. The first column displays the reference distributions used as ground truth. In the second column, \methodName\ successfully recovers the parameters in both cases with high fidelity, achieving a Mean Absolute Error (MAE) of $4.12 \times 10^{-6}$ in Case 1 and $2.76 \times 10^{-4}$ in Case 2. By contrast, the FNO baseline performs reliably only in Case 1 (MAE: $8.14 \times 10^{-4}$), where the geometry aligns with its training distribution of rectangles, triangles, and circles. Case 2 reveals a clear sensitivity to out-of-distribution morphologies; while the FNO correctly identifies the spatial position of the target, it erroneously predicts a shape resembling a training circle rather than the actual underlying geometry, resulting in an increased MAE of $1.26 \times 10^{-3}$. In the fourth column, PINN struggles significantly despite utilizing 50 times more data (50,000 vs. 1,000 measurements). In Case 1, PINN yields an MAE of $8.81 \times 10^{-3}$ which is about three orders of magnitude larger than \methodName\ and performs poorly overall. In Case 2, it fails to capture the random distribution of $D$ entirely, with the error further degrading to $3.55 \times 10^{-2}$.

Figure \ref{fig:2}d compares state variable $V$ predictions for Case 1 between \methodName\ and PINN. We note that FNO is excluded from this comparison because it cannot perform state variable predictions; unlike the physics-informed frameworks, FNO is trained as an end-to-end black-box estimator where the state variable serves as the input to predict the parameter. Furthermore, FNO is strictly constrained to structured grid data representations, whereas \methodName\ and PINN utilize flexible pointwise measurements. While PINN performs reasonably well within the training region (Fig. \ref{fig:2}d, first row), its peak error remains an order of magnitude higher than that of \methodName. For temporal extrapolation (second row), PINN exhibits significantly larger errors due to its inaccurate estimation of $D$. As shown in Fig. \ref{fig:2}a, \methodName\ consistently adheres to the underlying physics, providing robust predictions even beyond the training window. It achieves accurate temporal extrapolations beyond 40 $TU$ with an MAE of 0.00035 which is over two orders of magnitude lower than the PINN MAE of 0.066. Moreover, PINN frequently predicts non-physical negative $V$ values at multiple timesteps, further highlighting its limitations. A comprehensive analysis of these methodological differences and additional results under extremely sparse measurement conditions are provided in \textit{Supplementary Note 2} and \textit{2.2}.

\noindent\textbf{Implications:} The ability of \methodName\ to accurately estimate spatially heterogeneous parameters and predict state variables even under sparse, noisy conditions showcases its superiority over traditional methods like PINN. This advancement holds great promise for improving the diagnosis and modeling of cardiac pathologies, such as fibrosis and arrhythmias, where tissue heterogeneities play a critical role in understanding electrical signal propagation.

\begin{figure*}[!ht]
\centering
\includegraphics[width=\textwidth]{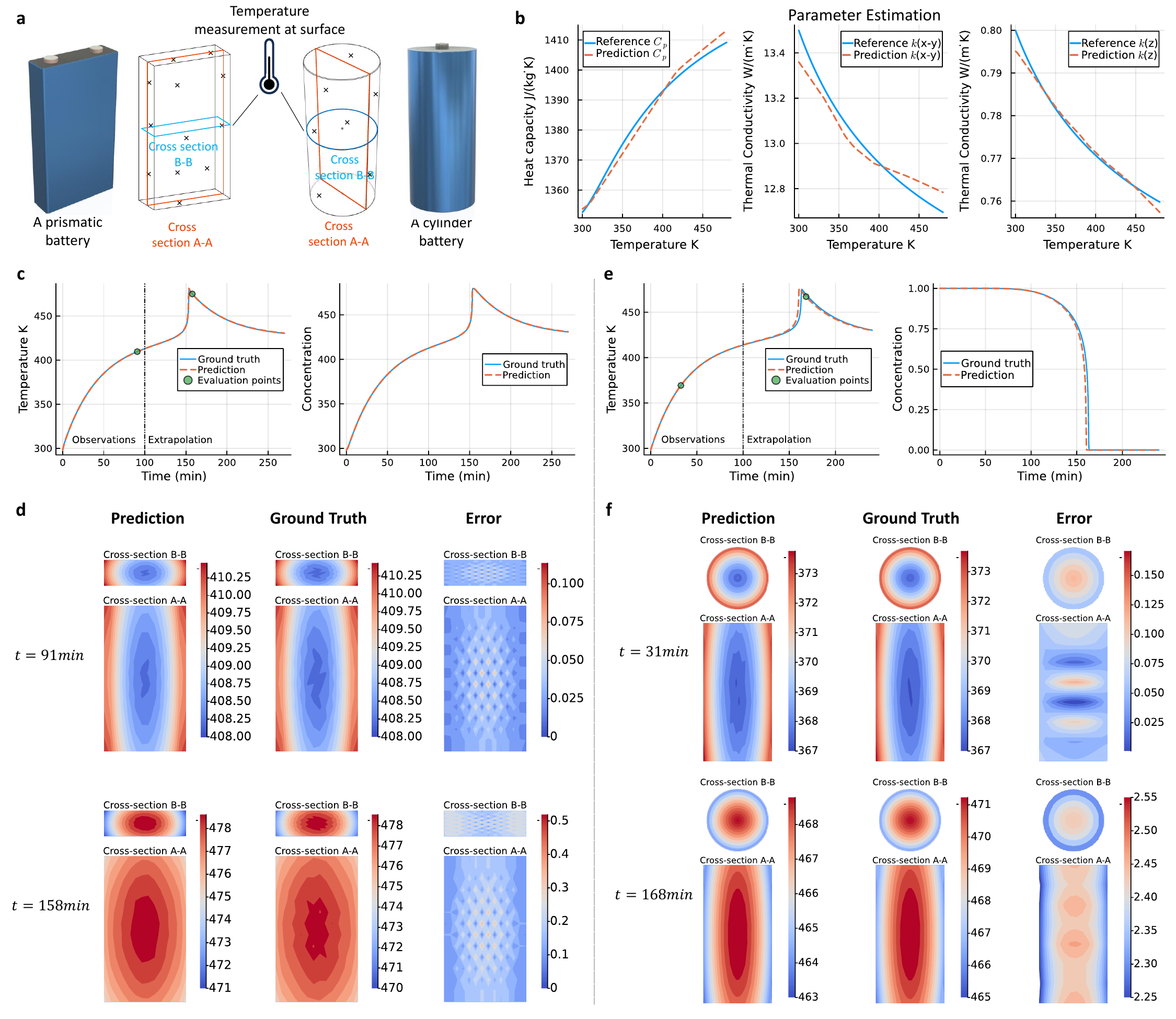}
\caption{\textbf{Parameter field estimation in thermal runaway problems.} \textbf{a}. A prismatic battery, a cylinder battery, and random temperature measurements at the battery surface. Cross-sections $A-A$ and $B-B$ represent the in-plane and the cross-plane directions, respectively. \textbf{b}. The estimated three parameter fields $C_p, k(x-y)$, and $k(z)$ for the prismatic battery. \textbf{c}. Predicted thermal runaway behaviors with the estimated parameters. The initial temperature of the battery is 25\textdegree$C$\ (or 298.15$K$), and the boundary condition (ambient temperature) is 423.15$K$. After accurately estimating the parameters, the model can correctly predict the thermal and chemical responses. \textbf{d}. Evaluation of temperature distribution for the prismatic battery at two typical times as marked in graph \textbf{c}. \textbf{e}. Predicted thermal runaway behaviors for the cylinder battery. \textbf{f}. Evaluation of temperature distribution for the cylinder battery. \label{fig:3}}
\end{figure*}

\section{Thermal Runaway in Batteries}
Modern lithium-ion batteries are designed to achieve high energy density, extended operational durations, and rapid charging capabilities, catering to the power requirements of contemporary electric vehicles (EVs). These high-performance batteries provide substantial benefits but also pose safety challenges, particularly the risk of thermal runaway (TR), which in extreme cases can result in fire accidents~\cite{hong2022investigation}. To address these risks, a thorough understanding of the underlying physics is crucial. 

\noindent\textbf{Known Physics:} TR is often governed by well-established thermal reactions and chemical kinetics, which are mathematically described by the following equations.
\begin{align} 
&\rho C_p \frac{\partial T}{\partial t} = \nabla ( k \nabla T) + \dot{Q}_{exo} \label{eq:thermal_PDE} \\ 
&\dot{Q}_{exo} = HW\left(-\frac{\partial c}{\partial t}\right) \label{eq:thermal_Q} \\
&\frac{\partial c}{\partial t} = -A_e exp\left(-\frac{E}{R_c T}\right)c \label{eq:thermal_c}
\end{align}

\noindent $T$ and $c$ represent the battery temperature and concentration of reacting materials. Other variables such as energy release $\dot{Q}_{exo}$, heat release $H$, active material content $W$, activation energy $E$, gas constant $R_c$, and porosity $\phi$ are described in more detail in \textit{Supplementary Note 1}.

\noindent\textbf{Unknown Parameters:} The parameters of interest ~\cite{loges2016thermal, werner2017thermal, loges2016study} are temperature-dependent heat capacity $C_p$, thermal conductivity $k$, and a scalar parameter called the pre-exponential factor $A_e$. For prismatic batteries, thermal conductivity is divided into cross-plane ($k_z$) and in-plane ($k_{xy}$) components, whereas for cylindrical batteries, it is characterized by radial $k_r$ and longitudinal $k_z$ components. The pre-exponential factor $A_e$ is specifically introduced to illustrate the generalizability of the method. 

These parameters are difficult to measure directly and are expected to behave as field variables influenced by the battery's internal states, such as temperature and other dynamically changing properties during operation. Additionally, properties like thermal conductivity and heat capacity can exhibit significant variability due to mass production inconsistencies and further evolve over their service life from aging effects~\cite{richter2017measurements} and environmental influences~\cite{an2019modeling}. This variability makes these parameters difficult to measure directly, but essential for understanding battery performance and predicting the likelihood of TR incidents.

\noindent\textbf{Problem Setup:} A standard prismatic battery ($100mm \times 180mm \times 32mm$) and a cylindrical battery ($66mm$ height, $28mm$ diameter) (see Fig. \ref{fig:3}a) are modeled to simulate TR behavior~\cite{wei2020comprehensive}. The initial condition is room temperature, with constant heat convection as the boundary condition (BC), and high external heat applied to simulate overheating.

\noindent\textbf{Measurements:} Considering real-life constraints, only surface temperatures can be measured, and only at limited locations. For training, we take just three measurement locations, with 15 temporal measurements each, during the initial 100 minutes before the irreversible temperature rise at around 150 minutes in TR. 

\noindent\textbf{Analysis:} Despite this limitation, \methodName\ accurately estimates the parameter fields ($C_p, k_{xy}$, and $k_z$). For the prismatic battery, the estimated parameters (Fig. \ref{fig:3}b) show MAE of 1.88, 0.052, and 0.001, compared to reference values of 1400, 13, and 0.8, respectively. For the cylindrical battery, the scalar parameter $A_e$ is jointly estimated, with an estimated value of $5.19\times 10^{25}$ showing a low percentage error of 0.97\% compared to the reference $A_e=5.4\times 10^{25}$. Similar accuracy is achieved for other parameters, with results detailed in \textit{Supplementary Note 2}. Furthermore, \methodName\ also demonstrates robustness under noisy measurements; for the prismatic battery, results for noise levels of 1\% and 10\% are detailed in Table 1.

\noindent\textbf{Implications:} Importantly, accurate parameter estimation enables the prediction of temperature changes of potentially ongoing TR. By integrating the parameters into the PDE system, predictions are extended over 0-270 minutes. Figure \ref{fig:3}c shows the temporal predictions of average temperature and concentration ($c$) profiles, closely matching the ground truth. Besides, inner temperature distributions can also be evaluated. In Fig. \ref{fig:3}d, representative results are displayed for cross-sections $A-A$ and $B-B$ (Fig. \ref{fig:3}a) at two key stages, before and during TR, where temperature profiles are accurately predicted with minimal errors.

Additionally, accurate parameter estimation using \methodName\ enables model customization to simulate battery behavior under diverse initial and boundary conditions (illustrated in \textit{Supplementary Note 2}). This capability supports predictive modeling for various scenarios, including environmental changes, manufacturing inconsistencies, and operational extremes, which are crucial for enhancing battery safety and performance. Unlike existing methods such as PINN, which often fail due to reliance on specific initial and boundary conditions, \methodName\ is independent of these constraints, ensuring consistent and reliable performance across scenarios. This robustness unlocks new opportunities for designing safer, more efficient batteries and evaluating their behavior over extended lifetimes or under extreme conditions.

\begin{figure*}[!ht]
\centering
\includegraphics[width=\textwidth]{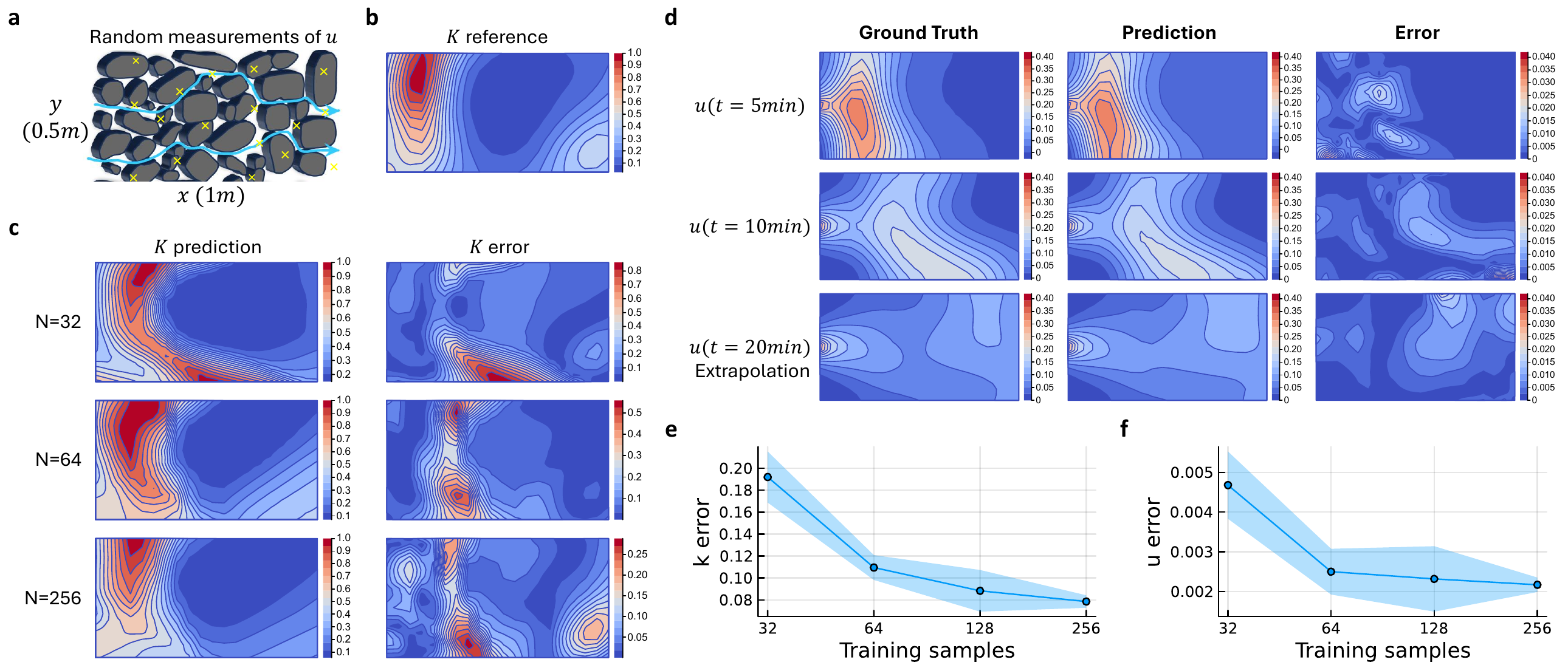}
\caption{\textbf{Parameter field estimation for flow in porous media.} \textbf{a}. A demonstration of flow phenomena, and random measurements are taken as the ground truth for model training. \textbf{b}. The estimated hydraulic conductivity $K$ compared to the reference $K$. \textbf{c}. The estimated parameter field $K$ and its estimation error compared to the reference, for different numbers of training samples. The magnitudes are normalized. \textbf{d}. The particle concentration predictions (state responses) at different times. The predictions align well with the ground truth, even for extrapolation at $t=20$ minutes. \textbf{e}. The parameter estimation performance compared to different numbers of training samples. Error bands are derived from standard deviations from multiple estimations and represent the variability in the performance. \textbf{f}. The response prediction performance compared to different numbers of training samples.\label{fig:4} }
\end{figure*}

\section{Flow in Porous Media}
The flow of solutes in porous media is a fundamental phenomenon observed in various natural systems~\cite{yan2022physics, yan2022gradient, rajabi2022analyzing, wen2021ccsnet, xu2023novel, du2023modeling, shokouhi2021physics}. This includes critical environmental concerns such as contaminant transport in water and the greenhouse effect caused by carbon emissions.

\noindent\textbf{Known Physics:} As an illustrative example, we focus on subsurface transport, a problem frequently studied in recent literature~\cite{he2021physics, he2020physics, tartakovsky2020physics}. This process is governed by coupled PDEs: the time-dependent advection-dispersion equation (ADE) and Darcy's law. These equations describe time-dependent transport behaviors and the physics of fluid flow in porous media, which are key to understanding and predicting such systems.
\begin{align}
&\frac{\partial u}{\partial t} + \nabla \cdot [\mathbf{v}u] = \nabla \cdot [\mathbf{D}\nabla u] \label{eq:flow-ade}\\
&\nabla \cdot [K \nabla h] = 0 \label{eq:flow-Darcy} \\
&\mathbf{v} = - K \nabla h / \phi \label{eq:flow-v}
\end{align}
\noindent where state variable $u$ represents the particle concentration field. Variables such as average pore velocity $\mathbf{v}$, dispersion coefficient $\mathbf{D}$, hydraulic head $h$, and porosity $\phi$ are described in more detail in \textit{Supplementary Note 1}.

\noindent\textbf{Unknown Parameters:} In this system, solute movement through porous media varies due to environmental conditions, human activities, and geological processes, making accurate parameter estimation of hydraulic conductivity $K(x,y)$ crucial for understanding these phenomena and addressing these pressing issues ~\cite{fienen2009obtaining, he2020physics, tartakovsky2020physics, wu2021physics}. However, $K$ is often difficult to measure directly, whereas particle concentration fields, depending on the system and measurement tools, can be relatively easier to retrieve.

\noindent\textbf{Problem Setup:} A 2D computational domain of $L_x=1m$ and $L_y=0.5m$ is defined \cite{he2021physics}, as shown in Fig.\ref{fig:4}a. Using just 32 random spatial-temporal concentration measurements $u$ over 0–16 minutes, the parameter field $K$ can be robustly estimated. Since the right-hand side of equation \ref{eq:flow-Darcy} is 0, multiple solutions exist, so the estimated $K$ is normalized and compared to the normalized reference (see Fig. \ref{fig:4}b).

\noindent\textbf{Measurements:} Figure \ref{fig:4}c shows the normalized $K$ estimation for different sample sizes ($N=$ 32, 64, 256). With fewer samples, the data may not fully capture the response, leading to larger prediction errors. However, even with just 32 points, \methodName\ already captures the key features of the parameter's distribution. Increasing the number of samples enhances the resolution of predictions and further reduces estimation errors.

\noindent\textbf{Analysis:} Figure \ref{fig:4}d illustrates the evolution of $u$ predictions over time. The first and second rows show contour predictions during the training region (at $t=5$ minutes and $t=10$ minutes) with low error and strong alignment with the ground truth. Notably, temporal extrapolation up to 20 minutes, beyond the 16-minute training region, accurately captures the concentration $u$, as shown in the third row of Fig. \ref{fig:4}d.

To demonstrate the robustness of \methodName, parameter estimation is repeated with varying sample sizes. Figure \ref{fig:4}e shows the parameter estimation performance, while Fig. \ref{fig:4}f displays the corresponding response prediction accuracy, with errors significantly reduced at 64 samples. Ablation studies on neural network sizes, detailed in \textit{Supplementary Note 4}, further validate the model's robustness and generalizability. For comparison with PINN-based approaches, we implemented state-of-the-art training techniques following recent best practices: residual-based adaptive sampling~\cite{lu2021deepxde,wu2023comprehensive} to dynamically refine collocation points in high-residual regions, 64-bit floating-point precision~\cite{xu2025fp64} to improve numerical stability, balanced residual decay rate (BRDR) loss weighting~\cite{chen2024self} to ensure balanced convergence across loss components, and a multi-phase training strategy with second-order optimization~\cite{wang2025gradient,urban2025unveiling} using L-BFGS-B for fine-tuning. The PINN implementation details and comparison results are provided in \textit{Supplementary Note 2.3}. Additionally, a spline-based parameter fitting method was tested as a non-neural network strategy but showed lower performance than \methodName\ {\color{black}(around 1.5 times }larger mean errors in both the parameter estimation and response prediction). The spline-based parameter fitting setups and results can be found in \textit{Supplementary Note 6}.

\noindent\textbf{Implications:} Accurately estimating the hydraulic conductivity field $K$ and predicting solute transport dynamics with \methodName\ enables robust modeling of subsurface transport, even with sparse and noisy measurements. This capability is critical for addressing environmental challenges such as contaminant transport and groundwater management, where accurate parameter estimation and temporal predictions are essential for effective decision-making.

\begin{figure*}[!ht]
\centering
\includegraphics[width=\textwidth]{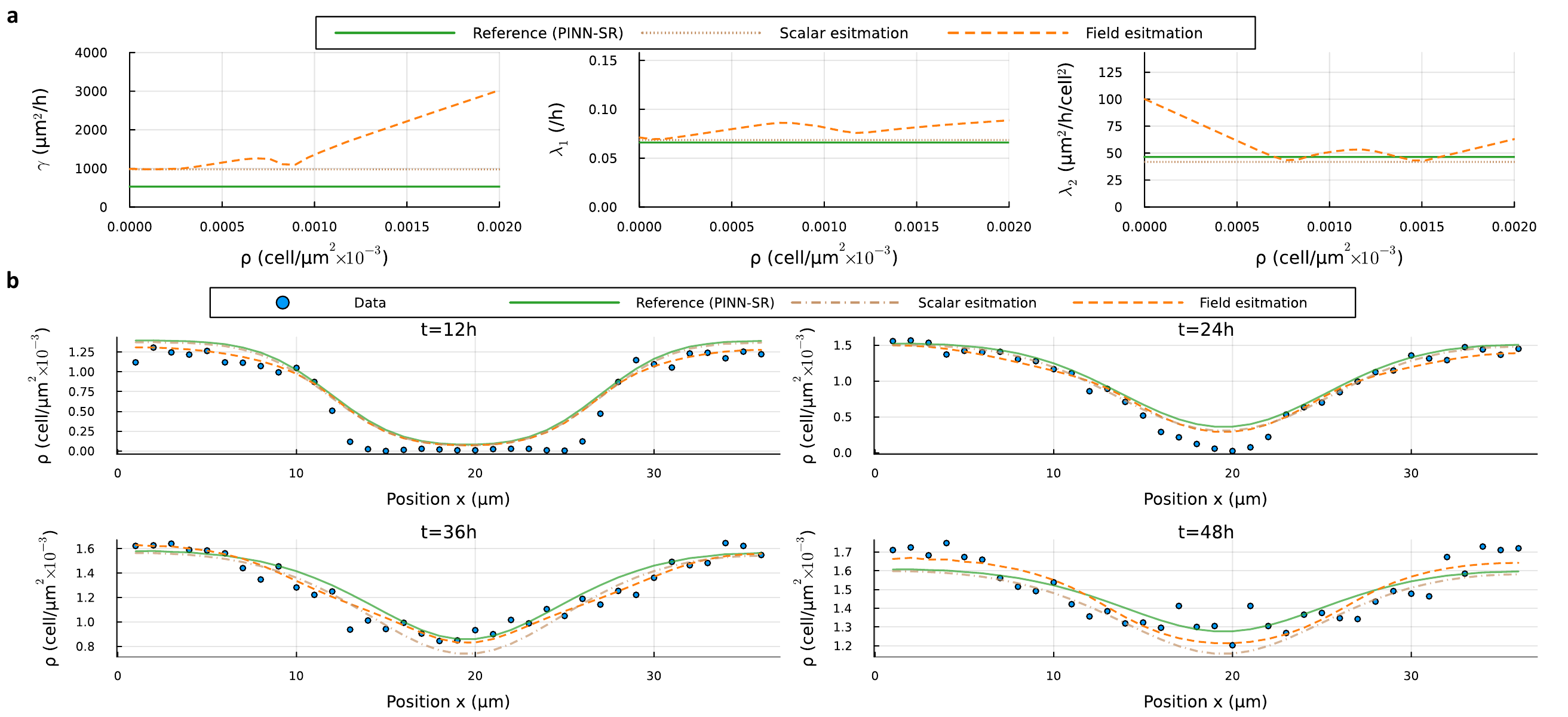}
\caption{\textbf{Parameter estimation in cell migration and proliferation.} \textbf{a}. Parameter estimation results and comparisons. \methodName\ assumes all three parameters are density-dependent. The results show an increasing $\gamma$ value as density increases, while the other two parameter estimations exhibit minimal variation and can be considered as constants. This aligns well with assumptions from previous research \cite{sengers2007experimental, vo2015quantifying}. \textbf{b}. Density response predictions. Using the estimated parameter fields, the density response at four different timesteps is displayed. Compared to results using the reference method (PINN-SR) and the scalar parameters setting, \methodName\ more accurately matches the experimental data at 36-hour and 48-hour, particularly in the regions with higher density values on both sides.}
\label{fig:5}
\end{figure*}

\section{Cell Migration and Proliferation}
\textbf{Application and Known Physics:} This section presents an experimental study of the diffusion (cell migration) and reaction (cell proliferation) process, which is governed by a known form of PDE \cite{fisher1937wave, maini2004traveling, chen2021physics}. 
\begin{equation}
\frac{\partial\rho}{\partial t} = \gamma\frac{\partial^2\rho}{\partial x^2} + \lambda_1 \rho - \lambda_2 \rho^2
\end{equation}
\noindent where parameters such as cell diffusivity $\gamma$ and proliferation rates $\lambda_1$ and $\lambda_2$ define the characteristics of cell migration and proliferation.

\noindent\textbf{Unknown Parameters:} These parameters are unknown and require estimation from cell density $\rho$ observations. According to the Porous-Fisher model~\cite{sengers2007experimental, vo2015quantifying}, cell diffusivity $\gamma$ varies with density (Fig. \ref{fig:5}a), increasing with higher cell density, while proliferation rates $\lambda_1$ and $\lambda_2$ are treated as constants. However, to demonstrate \methodName's generalizability, all 3 parameters are modeled as density-dependent fields. 

\noindent\textbf{Problem Setup:} {\color{black}Unlike previous sections which use synthetic data,} the experimental data here were originally extracted from high-resolution imaging collected in previous research on in vitro cell migration (scratch) assays \cite{jin2016reproducibility}, and were further analyzed \cite{chen2021physics} aimed at discovering the governing equation. The preprocessed data describes the cell density distributions at 38 evenly distributed spatial points in one dimension and only at 5 time steps ($0h, 12h, 24h, 36h, 48h$). The spatial domain ranges from 0 to 1900 $\mu m$. A detailed description of the experiment setup and data preprocessing can be found in the previous research \cite{jin2016reproducibility, chen2021physics}. This spatial-temporal experimental data is extremely sparse and noisy. 

\noindent\textbf{Measurements:} Given the initial condition at 38 points from the experimental data ($t=0h$) and Neumann boundary conditions $\partial\rho(x=0)/\partial x = 0$ and $\partial\rho(x=1900)/\partial x = 0$, we train \methodName\ with only 152 cell density data points over 38 spatial points and 4 timesteps. In the same setup, we compared our \methodName\ with PINN-SR \cite{chen2021physics} (i.e., PINN with sparse regression), for parameter estimation and response prediction performance. 

\noindent\textbf{Analysis:} As shown in the first graph of Fig. \ref{fig:5}a, \methodName\ estimates an increasing $\gamma$ with cell density, consistent with previous research~\cite{fisher1937wave, maini2004traveling}. Field estimations also suggest $\lambda_1$ and $\lambda_2$ are likely scalars. 
We also assume space-dependent parameters for estimation: results showing parameter distributions with high (peak) values at two boundaries, demonstrating that the parameter fields are density (or state) dependent. In the third graph, while the field estimation for $\lambda_2$ shows large values at low densities and stabilizes as density increases, this behavior is primarily due to highly noisy experimental data in regions with low cell numbers, rather than being attributed to the nature of the parameter field itself. In contrast, PINN-SR, limited by its methodology, cannot estimate parameter fields, and its scalar estimations show slightly different combinations but remain within a reasonable range. This limitation is evident at $12h$ (first graph) and $48h$ (last graph), where it fails to capture enhanced cell spreading.

Accurate parameter field estimation excels in predicting cell density distributions over time, as shown in Fig. \ref{fig:5}b. For such sparse and noisy data, PINN-SR achieves a median error of $6.53\times10^{-5}$. In contrast, \methodName, assuming density-dependent parameters, achieves the highest accuracy with an error of $4.71\times10^{-5}$. Across experimental data with varying initial cell densities~\cite{chen2021physics}, \methodName\ achieves 28-47\% lower median errors. Additional results are detailed in \textit{Supplementary Note 2}.

\noindent\textbf{Implications:} The findings demonstrate that \methodName\ not only provides more accurate parameter estimations under sparse and noisy conditions but also captures critical density-dependent dynamics that other methods, like PINN-SR, cannot. This capability enables improved predictions of biological phenomena such as cell migration and proliferation, offering new avenues for understanding and modeling reaction-diffusion systems in complex and noisy experimental scenarios.

{\color{black}
\section{Benchmarking and Ablation}
Having presented the four applications individually, we now summarize the comparison against baselines, the computational cost, and an ablation of the two-stage strategy.}

\noindent\textbf{Comparison with baselines.} 
Compared to sparse identification methods like PINN-SR, \methodName\ models full parameter field distributions rather than scalars, achieving 28–47\% improvement in forward response accuracy. This methodological shift enables more detailed modeling of the underlying physical processes, leading to improved predictive accuracy. Additionally, compared to PINN, our framework offers stronger temporal extrapolation, better adherence to physical laws, and greater adaptability to boundary and initial condition changes. 
It also requires less data, making it ideal when measurements are limited or costly. {\color{black}As detailed in Table \ref{tab:baseline_comparison}, which reports parameter estimation performance and computational load,  \methodName\ uses one to three orders of magnitude fewer measurements than PINN on the synthetic-data problems while matching or exceeding its accuracy. On porous flow it matches PINN using 64 measurements against 40,000, and on cardiac EP it reduces the parameter error by three orders of magnitude at one-fiftieth the data}.

\noindent\textbf{Computational cost.}
{\color{black}Table~\ref{tab:baseline_comparison} also reports iteration counts and wall-clock training time. 
\methodName\ converges in up to two orders of magnitude fewer iterations than the neural-network baselines, and does so entirely on CPU while the neural-network baselines utilize GPU acceleration. Iteration counts are those used to produce the reported results and follow standard settings for each architecture; they are not directly comparable across methods, since one \methodName\ iteration corresponds to a full forward PDE solve with adjoint backpropagation, whereas PINN and FNO iterations are minibatch updates. Wall-clock time is therefore the comparable measure. On this measure \methodName\ trains 3 to 20 times faster than physics-informed baselines across the four applications. 

Two entries warrant clarification. FNO's shorter training time for cardiac EP reflects GPU-accelerated supervised learning and excludes the cost of generating its 35,200 paired training samples, each of which requires a forward PDE solve. Spline fitting is {\color{black}slightly} faster than \methodName\ for porous flow but yields parameter errors around 1.5 times larger, as it fits the parameter field directly without solving the governing equations.
Both FNO and spline fitting also require gridded data and cannot operate directly on the scattered sparse measurements used by \methodName\ and PINN. We report training rather than inference time because, for inverse problems, the computational cost is dominated by the estimation stage; once trained, all methods produce predictions in negligible time, and the notion of inference differs across method types.}

\noindent\textbf{Two-stage versus one-stage estimation.}
{\color{black}The proposed two-stage learning strategy handles parameters across different scales by first estimating scalar magnitudes, allowing the neural network to focus on learning local variations within the parameter field. Table~\ref{tab:neptune_ablation} shows that its benefit lies primarily in optimization quality rather than wall-clock speed: end-to-end training time is comparable between the two variants in most applications (232 vs 233 min for battery; 47.9 vs 49.5 min for porous flow), with a clear speedup only for cell migration (4.5 vs 13.4 min), whereas the parameter error is lower by one to two orders of magnitude (1.88 vs 36.68 for battery; $4.12 \times 10^{-6}$ vs $1.17 \times 10^{-3}$ for cardiac EP).}  
Furthermore, we extend this discussion to demonstrate the superior performance of our strategy compared to other training approaches, including: (1) estimating only scalars, (2) directly estimating each parameter field in one step using separate neural networks; (3) estimating all parameter fields within a single neural network; and (4) jointly estimating scalar magnitudes and field variations in a single step. {\color{black}In particular, Table~\ref{tab:neptune_ablation} quantifies the comparison against the 1-stage variant, which directly optimizes the neural network field parameterization without the first-stage scalar estimation, across all four applications.} Detailed comparisons are provided in \textit{Supplementary Note 5}. 

\begin{table*}[!ht]
\centering
\caption{{\color{black}Comparison of Neptune against baseline methods across all applications, in terms of data requirements, training time, and parameter estimation accuracy (absolute error). Data Size indicates the number of spatial-temporal measurement points sampled for training, {\color{black}corresponding to the settings reported in the case studies above.} For FNO, Data Size is formatted as training/inference due to its supervised learning setup. For Battery Thermal, the two reported values correspond to $C_p$ and $k_{xy}$. For Cell Migration, real experimental data are used and ground-truth parameters are unavailable. Baseline neural-network methods (PINN, FNO, PINN-SR) utilize GPU acceleration (NVIDIA RTX A6000), {\color{black}whereas both Spline Fitting and Neptune run on a standard laptop CPU (Apple M1 Pro)}. For Porous Flow, errors are computed on normalized fields, as Darcy's law admits only relative recovery of $K$.}}\label{tab:baseline_comparison} 
\small
\begin{tabular}{llcccc}
\toprule
\textbf{Application} & \textbf{Method} & \textbf{Data Size} &
\textbf{Iterations} & \textbf{Training Time} &
\textbf{Parameter Error} \\
\midrule

\multirow{3}{*}{Cardiac EP}
& Neptune              & 1,000  & 4,000   & 312 min & $4.12\times10^{-6}$ \\
& PINN                 & 50,000 & 200,000 & 960 min   & $8.81\times10^{-3}$ \\
& FNO                  & 35,200/2,700  & 880,000     & 188 min   & $8.14\times10^{-4}$ \\
\midrule

\multirow{2}{*}{Battery Thermal}
& Neptune              & 45     & 2,250     & 232 min & 1.88,   $5.2\times10^{-2}$\\
& PINN                 & 40,000     &  70,000    &  711 min & 63.16, 1.92\\
\midrule

\multirow{3}{*}{Porous Flow}
& Neptune              & 64      & 3,600      & 47.9 min   &  $1.1\times10^{-1}$ \\
& PINN                 & 40,000     & 110,000      & 971 min   &  $1.2\times10^{-1}$ \\
& Spline Fitting       & 100     & 515      & 42 min   &  $1.6\times10^{-1}$\\
\midrule

\multirow{2}{*}{Cell Migration}
& Neptune              & 152     & 2,700      & 4.5 min   & -- \\
& PINN-SR              & 152     & 126,000      & 23.1 min   & -- \\
\bottomrule
\end{tabular}
\end{table*}

\begin{table*}[!ht]
\centering
\caption{{\color{black}Comparison of Neptune 1-stage and 2-stage variants in terms of data requirements, training time, and parameter estimation accuracy (absolute error). The 1-stage variant directly optimizes the neural network field parameterization without the first-stage scalar estimation. The 2-stage results are consistent with those reported in Table~\ref{tab:baseline_comparison}. All variants run on CPU (Apple M1 Pro). For Porous Flow, errors are computed on normalized fields, as Darcy's law admits only relative recovery of $K$. For Cell Migration, ground-truth parameters are unavailable and parameter error is not reported.}}\label{tab:neptune_ablation}
\small
\begin{tabular}{llcccc}
\toprule
\textbf{Application} & \textbf{Variant} & \textbf{Data Size} &
\textbf{Iterations} & \textbf{Training Time} &
\textbf{Parameter Error} \\
\midrule

\multirow{2}{*}{Cardiac EP}
& 2-stage & 1,000 & 4,000 & 312 min & $4.12\times10^{-6}$ \\
& 1-stage & 1,000 & 4,000 & 338 min   & $1.17\times10^{-3}$  \\
\midrule

\multirow{2}{*}{Battery Thermal}
& 2-stage & 45 & 2,250 & 232 min & 1.88,   $5.2\times10^{-2}$\\
& 1-stage & 45 & 2,860 & 233 min   & 36.68, $2.7\times10^{-1}$\\
\midrule

\multirow{2}{*}{Porous Flow}
& 2-stage & 64 & 3,600 & 47.9 min  &  $1.1\times10^{-1}$\\
& 1-stage & 64 & 3,600 & 49.5 min   & $1.8\times10^{-1}$ \\
\midrule

\multirow{2}{*}{Cell Migration}
& 2-stage & 152 & 2,700 & 4.5 min & -- \\
& 1-stage & 152 & 2,700 & 13.4 min & -- \\
\bottomrule
\end{tabular}
\end{table*}

\section{Discussion}
In this paper, we presented \methodName\ for accurate parameter field estimation under scarce observation data, demonstrated by various PDE problems in emerging fields where physical parameters are unknown and important. Specifically, we estimated the thermal conductivity and heat capacity in coupled thermal problems, enabling accurate TR prediction. In porous media flow, spatially varying hydraulic conductivity was recovered, improving the modeling of phenomena such as contaminant transport. For cardiac EP, tissue electrical conductivity was inferred solely from voltage measurements, offering potential for detecting fibrosis and other localized pathologies. Finally, the cell migration and proliferation case demonstrates effective parameter estimation under real-world noisy and sparse data conditions. It also highlights the method's capability in discovering governing equations and its generalizability in estimating scalar variables. 

In terms of utility, \methodName\ demonstrates its generalizability and robustness to estimate multiple parameters simultaneously across various dependencies (spatial and state), each exhibiting distinct patterns. {\color{black}This flexibility is further supported by the choice of discretization:} the use of FDM makes the method easily adaptable to a wide range of PDEs{\color{black}, including convection-dominated or smooth wave-type equations, provided a stable spatial discretization can be established. However, the reliance on FDM inherently restricts the framework to regular geometric domains and differentiable continuous fields, making it less effective for shock waves or large discontinuities. For the regular-domain problems studied here, however, it retains favorable computational efficiency and supports gradient-based optimization.}

However, inverse problems like parameter estimation remain ill-posed and prone to local minima, especially with complex fields or limited data. {\color{black} A related consideration is systems identifiability, an inherent property of the physical setup rather than the estimation algorithm. For elliptic and parabolic inverse problems such as those studied here, uniqueness results have been established under appropriate observation conditions~\cite{isakov2006inverse}, so the target coefficients are identifiable in principle, while their practical recovery depends on the amount and quality of data. The porous flow problem is a deliberate exception, where the homogeneous structure of Darcy's law admits only relative recovery of $K$, which we address by reporting normalized fields. In practice, identifiability degrades with sparser data or higher noise, consistent with the increasing errors in our sensitivity experiments (Fig.~\ref{fig:2}b, Fig.~\ref{fig:4}e--f, Table~\ref{tab:table2}).} These challenges can be mitigated by using informed initial guesses and incorporating prior knowledge, such as parameter magnitudes, smoothness penalties, and strategic data selection. Future work may extend the method to more general finite element frameworks {\color{black}to accommodate complex geometries and non-smooth dynamics, extend the empirical validation to other equation classes such as convection-dominated and hyperbolic systems, and explore its application to} chaotic systems.

\section{Methods}
\textbf{Problem Setup and differentiable PDE solver}
We consider a general time-dependent PDE of the form encompassing a range of PDE-governed systems studied in this work:
\begin{align} \label{eq:pde}
f\left( \frac{\partial \mathcal{S}}{\partial t}, \mathcal{S}, \nabla \mathcal{S}, \nabla^2 \mathcal{S}, t; \mathcal{P}, \ldots \right) = 0,
\end{align}
where $\mathcal{S}=\mathcal{S}(\mathcal{X}, t)$ is the state variable defined over spatial domain $\mathcal{X}$ (e.g., $x$, $y$, or $z$) and time interval $t \in [0, T]$. The unknown parameters $\mathcal{P}=[p_1,p_2,\dots]$ represent latent or field-dependent quantities of interest. The terms $\nabla \mathcal{S}$ and $\nabla^2 \mathcal{S}$ denote the spatial gradient and Laplacian of the state variable, respectively. \methodName\ directly solves PDE systems {\color{black}via the Method of Lines (MoL): FDM is applied for spatial discretization, reformulating the PDE into a semi-discrete ODE system, which is then integrated using differentiable ODE solvers.} This approach ensures efficient and accurate forward simulations while enabling gradient computation through the solver via the adjoint method. Using evenly spaced mesh nodes, the second derivative in the $x$-direction at a point $x=x_i$ is approximated by the central difference scheme~\cite{randall2005finite}:
\begin{equation}\label{eq:central-difference}
\frac{\partial^2 \mathcal{S}(\mathcal{X}_i, t)}{\partial \mathcal{X}^2} = \frac{\mathcal{S}_{i-1} - 2\mathcal{S}_i + \mathcal{S}_{i+1}}{\Delta \mathcal{X}^2} + \mathcal{O}(\Delta \mathcal{X}^2)
\end{equation}
\noindent where $\Delta \mathcal{X}$ is the mesh spacing and $\mathcal{S}_i$ is the discrete state value at $\mathcal{X}_i$. The term $\mathcal{O}(\Delta \mathcal{X}^2)$ denotes the truncation error, indicating second-order spatial accuracy.

In most scenarios considered in this study, the spatial dimensions of the state variables are organized into a 1D vector form for computational efficiency. The state variable and discretization matrices are structured as $u \in \mathbb{R}^{n \times 1}$ and $A_j \in \mathbb{R}^{n \times n}$, where $n = n_x \times n_y \times n_z$ denotes the total number of spatial nodes in the discretized 3D domain. Spatial derivatives of various orders are explicitly formulated for each direction, as shown in equation~\ref{eq:fdm}:
\begin{align} \label{eq:fdm}
\frac{\partial^{j} \mathcal{S}}{\partial \mathcal{X}^{j}} = A_{j} \mathcal{S},
\end{align}
\noindent where $j$ represents the derivative order and the matrix $A_j$ numerically approximates the corresponding spatial derivative.

Substituting the finite difference discretizations into the general PDE form yields the semi-discrete system:
\begin{align} \label{eq:discretized-pde}
f\left( \frac{d \mathcal{S}}{dt}, \mathcal{S}, A_1 \mathcal{S}, A_2 \mathcal{S}, t; \mathcal{P}, \ldots \right) = 0,
\end{align}
where $A_1$ and $A_2$ denote the numerical approximations of the first- and second-order spatial derivatives, respectively. Boundary conditions are carefully handled during discretization. Two types, Dirichlet and Neumann, are considered, expressed as $\mathcal{S}(\mathcal{X}) = \beta$ or $\frac{\partial \mathcal{S}(\mathcal{X})}{\partial \mathcal{X}} = \beta$, where $\beta$ is the prescribed coefficient. Boundary conditions are implemented by modifying the discretization matrices in equation~\ref{eq:fdm}, following standard approaches~\cite{leveque2007finite, li2022neuralsi}. After discretization, the PDE system is reformulated as a set of ODEs and solved using Runge–Kutta methods~\cite{dormand1980family}, with dynamics extracted at observation times.

\begin{figure*}[!ht]
\centering
\includegraphics[width=1\textwidth]{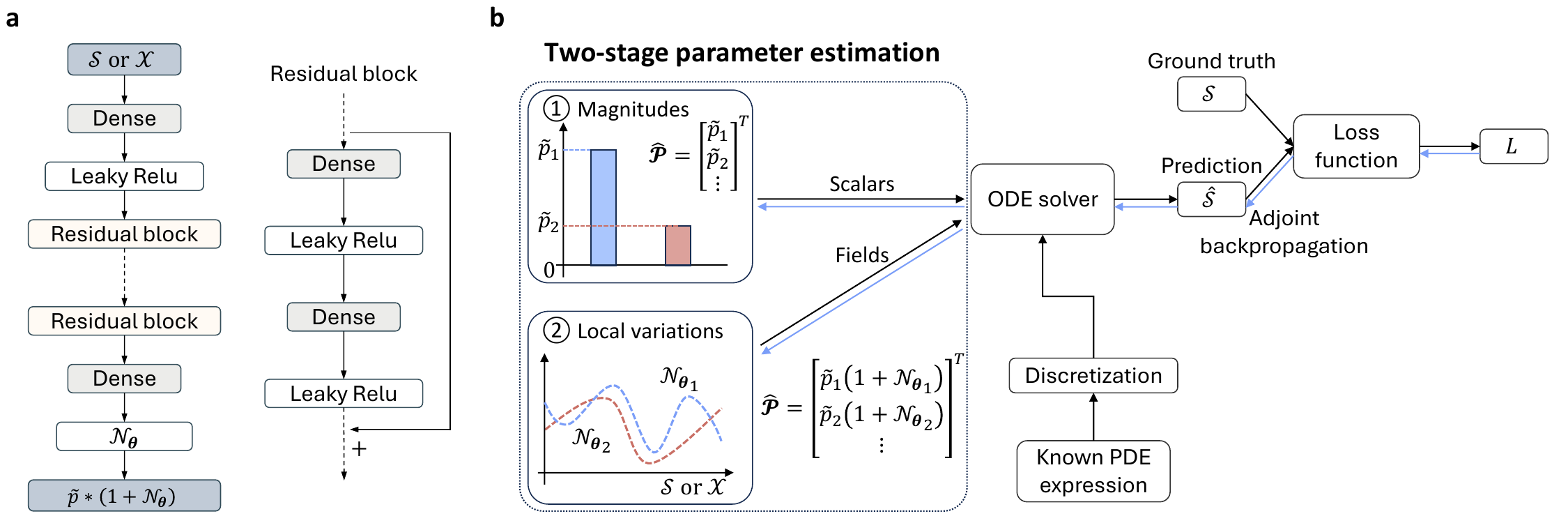}
\caption{\textbf{The schematic diagram of the two-stage parameter field estimation strategy and the neural network architecture.} \textbf{a}. The feed-forward neural network. Left: overall architecture. Right: residual block structure. For flow and cardiac electrophysiology problems, the input to the network is the spatial coordinate $\mathcal{X}$. For battery thermal runaway and cell migration, the input is the state variable $\mathcal{S}$ (e.g., $T$ or $\rho$), with fewer layers used. The network output $\mathcal{N}_{{\boldsymbol{\theta}}}$ is scaled to the physical parameter as $\tilde{\mathcal{P}}\big(1+\mathcal{N}_{{\boldsymbol{\theta}}}\big)$. The reference value $\tilde{\mathcal{P}}$ is application-specific and determined via a two-step training strategy. \textbf{b}, Parameter estimation diagram. The left part illustrates the two-stage estimation for scalar and parameter fields, while the right part shows the numerical solving and optimization procedure.\label{fig:6}}
\end{figure*}

\textbf{Parameter field modeling with neural networks} 
For multiple unknown parameter fields $\mathcal{P}=[p_1, p_2, \dots]$, \methodName\ simply employs different scalars $\tilde{p}_i$ and networks $\mathcal{N}_{\boldsymbol{\theta}_i}$ for parallel modeling, where $i=1,2,\dots$. Specifically, each parameter field $p_i$ is modeled as:
\begin{align}
&p_i = \tilde{p}_i\big(1+\mathcal{N}_{\boldsymbol{\theta}_i}(\mathcal{X} \text{ or } \mathcal{S})\big) \label{eq:nn+1} \\
&\mathcal{N}_{\boldsymbol{\theta}_i}= f_{m} \circ f_{m-1} \circ \cdots \circ f_{1}(\mathcal{X} \text{ or } \mathcal{S})
\label{eq:net}
\end{align}
\noindent where $\tilde{p}_i$ denotes a learnable scalar and $\mathcal{N}_{\boldsymbol{\theta}_i}$ represents the neural network with $m$ layers and learnable weights as $\boldsymbol{\theta}_i$. A detailed architecture with skip connections can be referred to Fig. \ref{fig:6}. $f$ denotes a neural layer and $\circ$ represents function composition. 
{\color{black}The input to $\mathcal{N}_{\boldsymbol{\theta}_i}$ depends on the physical characteristics of the parameter field: it takes discretized spatial coordinates $\mathcal{X}$ (e.g., $x, y$) for the porous media flow and cardiac electrophysiology problems, or state variables $\mathcal{S}$ (e.g., temperature $T$ or cell density $\rho$) for the battery thermal runaway and cell migration problems. These inputs are properly scaled (i.e., -1 to 1 from $\mathcal{X}$ or observed $\mathcal{S}$).}

{\color{black}Because the neural network is evaluated at each time step during ODE integration, the framework naturally accommodates state-dependent parameters that evolve implicitly over time (e.g., $k(T)$ in the battery problem); explicitly time-dependent fields can be accommodated by including $t$ as an additional network input.} 
Leaky ReLU activation functions are used in all layers except the final one, which omits activation functions. 

\textbf{Two-stage parameter estimation strategy}
\methodName\ employs a two-stage learning, first optimizing $\tilde{\mathcal{P}}$ and then training $\mathcal{N}_{\boldsymbol{\theta}}$. In the first stage (see the upper row of Fig. \ref{fig:6}), random scalars $\tilde{\mathcal{P}}$ are initialized {\color{black}from a uniform distribution spanning a factor of three above and below typical reference values, and optimized directly in their physical scales using per-parameter learning rates so that parameters differing by orders of magnitude receive comparable relative updates. The scalars are} incorporated into the discretized governing equations to numerically solve for the dynamic response. This is equivalent to setting $\boldsymbol{\theta}=\mathbf{0}$ in equation \ref{eq:nn+1}. Those learnable scalars are optimized using the adjoint method~\cite{chen2018neural} (detailed later in the section) by minimizing the error between $\hat{\mathcal{S}}$ and the observed measurements $\mathcal{S}$. The second stage fixes the converged scalars and only optimizes the neural network weights $\boldsymbol{\theta}$, aiming to estimate local variations in a reduced search domain (see the lower row of Fig. \ref{fig:6}). 

During the optimization, scalars $\tilde{\mathcal{P}}$ act as a physics-based prior and implicit regularizer, ensuring early physical stability of training and providing a well-scaled baseline for convenient expansion to parameter fields in the second stage. Even if $\tilde{\mathcal{P}}$ converges with high error, its combination with the PDE remains physically meaningful and stable during optimization.

In the second stage, the neural network acts primarily as a correction term, analogous to residual learning architectures such as ResNet~\cite{he2016deep}. Initially, the fixed scalar $\tilde{\mathcal{P}}$ constrains the neural network to explore only perturbations around a physically meaningful baseline. This constraint stabilizes gradient flow, prevents nonphysical field magnitudes, preserves the well-posedness of the underlying PDE, and generally improves training stability. Meanwhile, in later stages of training, the neural network is allowed unbounded exploration (in most cases, no activation is applied to the final layer), enabling it to capture more complex behaviors beyond the initial physics-based prior.

{\color{black}Compared to directly learning the parameter field with a neural network from the start, the first-stage scalar learning reduces the per-iteration cost, as solving the PDE with a full field parameterization is more expensive; its primary benefit, however, lies in optimization stability and final accuracy rather than wall-clock speed.}

In addition, by anchoring the field prediction around a physically meaningful baseline, this strategy improves generalization and robustness during training. The gradient flow with respect to $\boldsymbol{\theta}$ inherits better conditioning from the scalar optimization step, further enhancing convergence efficiency and training stability. Mathematically, the gradient with respect to the scalar parameter in the first stage is given by
\begin{equation}
\frac{dL}{d\tilde{\mathcal{P}}} = \frac{dL}{d\mathcal{P}} \frac{d\mathcal{P}}{d\tilde{\mathcal{P}}} = \frac{dL}{d\mathcal{P}}, \label{eq:gradient-scalar}
\end{equation}
where $d\mathcal{P}/d\tilde{\mathcal{P}}=1$ because $\mathcal{P}=\tilde{\mathcal{P}}$ in stage one.

In the second stage, following the reparameterization in equation \ref{eq:nn+1}, the gradient with respect to the network parameters $\boldsymbol{\theta}$ becomes
\begin{equation}
\frac{dL}{d\boldsymbol{\theta}} = \frac{dL}{d\mathcal{P}} \frac{d\mathcal{P}}{d\mathcal{N}_{\boldsymbol{\theta}}}\frac{d\mathcal{N}_{\boldsymbol{\theta}}}{d\boldsymbol{\theta}} = \tilde{\mathcal{P}} \frac{dL}{d\mathcal{P}} \frac{d\mathcal{N}_{\boldsymbol{\theta}}}{d\boldsymbol{\theta}}.  \label{eq:gradient-nn}
\end{equation}
Thus, the core physical gradient $\frac{dL}{d\mathcal{P}}$ is consistently inherited across both stages, ensuring smooth convergence and stable optimization dynamics.

\textbf{Neural network architecture and problem-specific design}
The coordinate neural networks (feed-forward) in the second stage are adaptively designed, with the number of residual blocks and neurons per layer customized for each application. Specifically, in the thermal battery problem, the scalar references for thermal conductivity in different directions ($\tilde{k}_{xy}$, $\tilde{k}_z$) and heat capacity ($\tilde{C}_p$) are first determined. 
In the cell migration problem, three scalar parameters, $\tilde{\gamma}$, $\tilde{\lambda_1}$, and $\tilde{\lambda_2}$, are similarly estimated.
In both cases, the neural network models state-dependent parameter fields based on a single state variable, with a network input size of 1 (e.g., temperature $T$ or density $\rho$, rescaled approximately from -1 to 1).
A compact network with 20 neurons per hidden layer and two residual blocks is used to ensure efficient training.

In contrast, for the flow and cardiac applications, the conductivity parameters $K$ or $D$ depend on a 2D space, requiring larger neural networks: 50 neurons with 3 residual blocks for the flow problem, and 150 neurons with 4 blocks for the cardiac problem. The spatial inputs ($x$, $y$) are rescaled from $-1$ to $1$ in both dimensions. Notably, in the flow problem, multiple parameter distributions can satisfy the same setup due to the nature of equation \ref{eq:flow-Darcy}, which involves taking spatial derivatives. Because its absolute magnitude is mathematically unidentifiable, the first-stage scalar optimization simply anchors a uniform macroscopic scale, allowing the second-stage network to efficiently learn the relative spatial variations.
Since scaling $K$ by any constant leaves the equation unchanged, $K$ is reported after normalization. An absolute value activation with a small offset ($1\times10^{-3}$) is applied at the final layer to ensure strictly positive predictions and avoid physically meaningless zeros.

Additionally, \textit{Supplementary Note 4} presents an ablation study evaluating the impact of neural network size on parameter estimation performance, while \textit{Supplementary Note 5} provides a detailed comparison of this training strategy against alternatives, including direct parameter field learning and using a single neural network for all parameters.

\textbf{Loss function and model training} \methodName\ relies solely on observations from physical phenomena for training. The loss function is defined purely using the mean absolute error (MAE):
\begin{equation}
L = \frac{1}{N} \sum_{i=1}^{N} |\mathcal{S}_i - \hat{\mathcal{S}}_i|
\end{equation}
\noindent where $\mathcal{S}$ and $\hat{\mathcal{S}}$ represent the ground truth (observations) and predicted responses, respectively, $N$ is the number of points. Preliminary experiments showed that mean squared error (MSE) yielded similar performance, and thus, MAE was chosen for simplicity. Regularization terms (e.g., weight decay) were also tested but adversely affected optimization and were therefore omitted.

During training, observations are split into mini-batches of size 16 (reduced to 8 for the cell migration problem due to limited samples). In the first stage, the loss is minimized to directly learn $\tilde{\mathcal{P}}$, with the learning rate adaptively decreasing until convergence, where $\tilde{\mathcal{P}}$ stabilizes without further change. Dashed lines in the figure illustrate the gradient flow during optimization.
In the second stage, the loss is minimized to optimize the neural network weights $\boldsymbol{\theta}$, with a smaller learning rate ranging from $10^{-4}$ to $10^{-2}$ depending on network size. The best-performing $\boldsymbol{\theta}$, corresponding to the lowest loss value, is saved as the final trained model. The parameter field is then predicted using equation \ref{eq:nn+1} with the corresponding spatial or state inputs.

To update the learnable weights ($\tilde{\mathcal{P}}$ or $\boldsymbol{\theta}$), \methodName\ employs gradient-based optimization through adjoint sensitivity analysis~\cite{chen2018neural}, {\color{black}which leverages advances in neural ODEs and reverse-mode automatic differentiation~\cite{rackauckas2020universal, rackauckas2019diffeqflux} for efficient gradient computation. Sampling-based alternatives such as Monte Carlo methods are computationally impractical for the high-dimensional parameter fields considered here~\cite{stuart2010inverse}.}
For the scalar parameters $\tilde{p}$, the adjoint method directly computes the loss gradient through the PDE solver, as shown in equation \ref{eq:gradient-scalar}. For the neural network weights $\boldsymbol{\theta}$, an additional application of the chain rule through the neural network is required, as detailed in equation \ref{eq:gradient-nn}.

Importantly, \methodName\ differs fundamentally from traditional neural ODE approaches: instead of modeling unknown or partially known governing expressions with neural networks, it tackles complex PDEs with fully known equations, while neural networks serve only as correction terms for modeling unknown parameter fields.
To clarify the underlying optimization, we next emphasize the derivation of the common term $\frac{dL}{d\mathcal{P}}$ for a standard differential equation. The adjoint method requires defining a scalar functional $g(\mathcal{S}, \mathcal{P})$, where $\mathcal{S}(t, \mathcal{P})$ is the numerical solution to the differential equation $\frac{d\mathcal{S}(t, \mathcal{P})}{dt} = f(t, \mathcal{S}, \mathcal{P})$ with $t \in [0, T]$ and initial condition $\mathcal{S}(t_0, \mathcal{P}) = \mathcal{S}_0$. In our experiments, we assume a known initial condition. 

In our work, though the PDE system is discretized spatially using equation \ref{eq:discretized-pde}, this semi-discretization does not fundamentally alter the adjoint sensitivity formulation. Adjoint sensitivity analysis is still applied to compute the gradient of $g$ with respect to the parameters $\mathcal{P}$:
\begin{equation}
    L(\mathcal{S}, \mathcal{P}) = \int_{t_0}^{T} g(\mathcal{S}(t, \mathcal{P}), \mathcal{P}) \, dt\label{eq:dLdtheta}
\end{equation}

To derive the adjoint equation~\cite{cao2003adjoint}, we introduce the Lagrange multiplier $\boldsymbol{\lambda}$ to form:
\begin{equation}
I(\mathcal{P}) = L(\mathcal{P}) - \int_{t_0}^{T} \boldsymbol{\lambda}^\top \left( \frac{d\mathcal{S}}{dt} - f(t, \mathcal{S}, \mathcal{P}) \right) \, dt
\end{equation}
Since $\mathcal{S}' = f(\mathcal{S}, \mathcal{P}, t)$, we have that:
\begin{align}
\frac{dL}{d\mathcal{P}} = \frac{dI}{d\mathcal{P}} &= \int_{t_0}^{T} \left( \frac{\partial g}{\partial \mathcal{P}} + \frac{\partial g}{\partial \mathcal{S}}\frac{\partial \mathcal{S}}{\partial \mathcal{P}} \right) dt \notag \\
&\quad - \int_{t_0}^{T} \boldsymbol{\lambda}^\top \left( \frac{d}{dt} \left( \frac{\partial \mathcal{S}}{\partial \mathcal{P}} \right) - \frac{\partial f}{\partial \mathcal{S}} \frac{\partial \mathcal{S}}{\partial \mathcal{P}} - \frac{\partial f}{\partial \mathcal{P}} \right) dt,
\end{align}
\noindent where $\frac{\partial \mathcal{S}}{\partial \mathcal{P}}$ represents the sensitivity of the state variables to the parameters. After applying integration by parts to the term involving $\boldsymbol{\lambda}^\top \frac{d}{dt}\left(\frac{\partial \mathcal{S}}{\partial \mathcal{P}}\right)$, we require that:

\begin{align}
\frac{d\boldsymbol{\lambda}(t)}{dt} = 
\frac{\partial g(\mathcal{S}(t, \mathcal{P}), \mathcal{P})}{\partial \mathcal{S}} 
&- \boldsymbol{\lambda}(t) \frac{\partial f(t, \mathcal{S}(t, \mathcal{P}), \mathcal{P})}{\partial \mathcal{S}}, \notag \\
\boldsymbol{\lambda}(T) &= 0
\end{align}

\noindent where $\frac{\partial f}{\partial\mathcal{S}}$ is the Jacobian of the system with respect to the state $\mathcal{S}$, and $\frac{\partial f}{\partial\mathcal{P}}$ is the Jacobian with respect to the parameters. 
{\color{black}To handle coupled multiphysics systems, such as the thermal runaway model, the discretized state variables are concatenated into an augmented vector (e.g., $\mathcal{S} = [T_1, ..., T_n, c_1, ..., c_n]^\top$). The cross-dependencies between physical fields are inherently captured by the off-diagonal blocks of the system's Jacobian $\frac{\partial f}{\partial \mathcal{S}}$. For example, in the thermal runaway case, this matrix takes the form 
$\frac{\partial \mathbf{f}}{\partial \mathcal{S}} = \begin{bmatrix} \frac{\partial \dot{\mathbf{T}}}{\partial \mathbf{T}} & \frac{\partial \dot{\mathbf{T}}}{\partial \mathbf{c}} \\ \frac{\partial \dot{\mathbf{c}}}{\partial \mathbf{T}} & \frac{\partial \dot{\mathbf{c}}}{\partial \mathbf{c}} \end{bmatrix}$, 
where the non-zero off-diagonal blocks, such as $\frac{\partial \dot{\mathbf{T}}}{\partial \mathbf{c}}$, encode the physical interactions between temperature and chemical concentration. The porous flow problem involves a different coupling pattern: Darcy's equation is a steady-state constraint, so the hydraulic head and velocity field are first computed from the current $K$ at each forward evaluation and then substituted into the time-dependent advection-dispersion equation, with gradients propagating through this sequential solve via the same automatic differentiation. In practice, these coupled interactions are automatically preserved via automatic differentiation in Julia, without requiring problem-specific adjoint derivations for each multiphysics configuration.}

The solution to the adjoint problem provides the sensitivities through the integral:
\begin{equation}
\frac{dL}{d\mathcal{P}} = \int_{t_0}^{T} \left( \boldsymbol{\lambda}^\top \frac{\partial f}{\partial \mathcal{P}} + \frac{\partial g}{\partial \mathcal{P}} \right) dt + \boldsymbol{\lambda}^\top(t_0) \frac{\partial \mathcal{S}(t_0)}{\partial \mathcal{P}}
\end{equation}

To compute the term $ \frac{\partial L}{\partial \mathcal{P}}$, the adjoint problem is formulated and solved as an ODE system. This is achieved using a specialized sensitivity analysis framework \cite{rackauckas2020universal}, which provides a range of numerical methods tailored to different types of PDE systems.\\
\textbf{How \methodName\ outperforms PINN in inverse parameter estimation tasks?} 
Here we demonstrate why PINN-based inverse parameter estimation fails and how our adjoint-based method fundamentally differs in its approach. This analysis uses flow in porous media as a representative example, though the moving target problem arising from simultaneous updates of parameters and state variables is general and applicable to other inverse problems where parameters $\mathcal{P}$ and state variables $\mathcal{S}$ are coupled through PDE constraints. 

In the baseline PINN formulation, the total loss 
$L_{\text{PINN}}$ consists of several components: 
$L_{\text{PDE}}$ represents the PDE residual losses enforcing the governing equations; 
$L_{\text{BC}_i}$ represent boundary condition losses (8 terms in total) enforcing Dirichlet and Neumann conditions on the domain boundaries. 
These boundary conditions are defined as follows:
Hydraulic head ($h$):
(1) $x=0$: Neumann flux $q=1$ (inflow); 
(2) $x=L_x$: Dirichlet $h=0$ (fixed head); 
(3) $y=0$: Neumann no-flux; 
(4) $y=L_y$: Neumann no-flux.
Concentration ($u$):
(5) $x=0$: Dirichlet Gaussian injection 
$u=\exp\left(-1600(y-L_y/2)^2\right)$; 
(6) $x=L_x$: Neumann zero-gradient outflow; 
(7) $y=0$: Neumann no-flux; 
(8) $y=L_y$: Neumann no-flux. $L_{\text{IC}}$ is the initial condition loss at $t=0$; 
and $L_{\text{data}}$ is the data mismatch term comparing predicted values to observational data.
The loss is a weighted sum: 
\begin{equation}
L_{\text{PINN}} = w_1 L_{\text{PDE}} + \sum_{i=1}^{8} w_{i+1} L_{\text{BC}_i} + w_{10} L_{\text{IC}} + w_{11} L_{\text{data}}
\end{equation}
where $w_i$ are loss weights. 

The inverse estimation of the parameter field $\mathcal{P}$ faces a fundamental optimization challenge arising from the coupled nature of the PDE residuals. The parameter $\mathcal{P}$ (represented by neural network weights $\boldsymbol{\theta}$) is updated exclusively through the gradients of the PDE residuals, which are the only non-zero terms in $\frac{\partial L_{\text{PINN}}}{\partial \boldsymbol{\theta}}$. 
However, these same loss terms simultaneously update the state variables $\mathcal{S}$ through their respective gradients $\frac{\partial L_{\text{PINN}}}{\partial \mathcal{S}}$. The coupling is evident from the residual dependency $R = R(\mathcal{S}, \mathcal{P})$. This creates a moving target problem during optimization: at iteration $t$, the parameter $\mathcal{P}^{(t)}$ attempts to minimize the PDE residuals $R(\mathcal{S}^{(t)}, \mathcal{P}^{(t)})$ based on the current state $\mathcal{S}^{(t)}$, but these state variables are concurrently being updated to $\mathcal{S}^{(t+1)}$ by the same residual terms. As a result, the optimization landscape for $\mathcal{P}$ shifts with each update, preventing stable convergence to the true parameter values.
In contrast to the PINN approach, \methodName\ circumvents this optimization conflict by decoupling parameter estimation from state variable learning. The forward problem is solved using a numerical solver, which enforces a deterministic relationship $\mathcal{S} = \text{Solver}(\mathcal{P})$ between the parameter and the state variable. Critically, during optimization, only the neural network weights $\boldsymbol{\theta}$ are updated; the state variable $\mathcal{S}$ is not a free optimization variable but is uniquely determined by $\mathcal{P}$ through the differentiable PDE solver at each iteration. This direct, one-way coupling eliminates the moving target problem: at iteration $t$, the parameter $\mathcal{P}^{(t)}$ minimizes the loss based on $\mathcal{S}^{(t)}$, and the subsequent state $\mathcal{S}^{(t+1)}$ is fully determined by the updated parameter, ensuring a stable optimization landscape that enables reliable convergence to the true parameter values.

\textbf{Data generation} 
For the cell migration and proliferation problem, experimental data are sourced from previous research~\cite{jin2016reproducibility}. For other data, we generate data by employing FDM to solve the PDEs in each problem. For battery thermal runaway modeling, the mesh size is set to $8$ in each spatial dimension, with a timestep of $200\,\text{s}$ (i.e., $100$ minutes over $30$ steps). {\color{black}For the cylindrical battery, we exploit axial symmetry to reduce the problem to a 2D $r-z$ plane, where the governing equations are formulated in cylindrical coordinates and the circular cross-section maps onto a regular rectangular grid for standard FDM discretization.} The two state variables, temperature $T$ and concentration $c$, are updated simultaneously at each step, along with the temperature-dependent parameters.
Secondly, in the flow problem, the PDEs are solved in multiple stages. The mesh size is set to $22 \times 22$, and a reference parameter $K$, based on previous research~\cite{he2020physics}, is regenerated. The hydraulic head $h$ is computed using Darcy's law (equation~\ref{eq:flow-Darcy}), followed by the velocity fields in the $x$ and $y$ directions, solved using $K$ and $h$ (equation~\ref{eq:flow-v}). 
{\color{black}Specifically, the term $\nabla\cdot(K\nabla h)$ is expanded using the product rule and discretized as $\mathbf{K} \odot (\mathbf{A}_2 \mathbf{h}) + (\mathbf{A}_{1x}\mathbf{K}) \odot (\mathbf{A}_{1x}\mathbf{h}) + (\mathbf{A}_{1y}\mathbf{K}) \odot (\mathbf{A}_{1y}\mathbf{h}) = \mathbf{b}$, where $\mathbf{A}_{1x}, \mathbf{A}_{1y}$ are the first-order finite-difference matrices in the $x$ and $y$ directions, $\mathbf{A}_2 = \mathbf{A}_{2x} + \mathbf{A}_{2y}$ is the second-order matrix including boundary modifications that depend on $\mathbf{K}$, $\odot$ denotes element-wise multiplication, and $\mathbf{b}$ collects the boundary contributions (see \textit{Supplementary Note 3} for detailed matrix formulations). The discrete gradients $\mathbf{A}_{1x}\mathbf{K}$ and $\mathbf{A}_{1y}\mathbf{K}$ therefore enter the operator explicitly.}
The resulting velocity field is then used to solve the time-dependent concentration equation for $u$ (equation~\ref{eq:flow-ade}).
Thirdly, in the cardiac electrophysiology problem, the mesh size is set to $50$ in both $x$ and $y$ dimensions. The two state variables, $V$ and $W$, are computed at each timestep (1$TU$) during the simulation.
Lastly, in the cell migration and proliferation problem, the mesh size is fixed at $38$ points, corresponding to the spatial measurement positions. Due to varying gaps between the first two and the last two points, the standard FDM is modified to accommodate a non-uniform finite difference mesh. There are $5$ timesteps: the first for setting initial conditions, and the remaining $4$ for training.

Notably, \textit{Supplementary Note 3} expands on the mathematical derivation of finite difference discretization~\cite{liszka1980finite, chung1981generalized, trew2005generalized}, explains the treatment of non-uniform meshes at geometry boundaries for the cell migration and proliferation problem, and discusses limitations of FDM. The training data, consisting of only the state responses, is generated numerically using the PDE solver for the first three scenarios, with different levels of Gaussian noise added to ensure robustness (details in \textit{Supplementary Note 1}).

\textbf{Data availability}
All synthetic training data are generated on-the-fly by our provided scripts. The experimental cell migration data, originally published in~\cite{jin2016reproducibility}, are also packaged within our code repository. Additional information can be found in the \textit{Code availability} section.

\textbf{Code availability}
The source code used to generate the results presented in this study is provided as \textit{supplementary file} for the peer-review process. The main algorithm is implemented in Julia, while baseline methods including PINNs and neural operators are implemented in Python (PyTorch). Upon acceptance, the code will be hosted on a public GitHub repository.

\textbf{Author contributions} N.L. and V.B. supervised the study. X.L., N.L., and V.B. conceived the initial idea. X.L. and M.M. conducted the numerical experiments. X.L., M.M., N.L., and V.B. wrote the paper. X.L., M.M., and R.G. contributed to the supplementary.

\textbf{Competing interests} The authors declare no competing interests.

\textbf{Correspondence and requests for materials} should be addressed to \href{mailto:vishnu@msu.edu}{vishnu@msu.edu} and \href{mailto:lajnefni@msu.edu}{lajnefni@msu.edu}.


\clearpage
\onecolumngrid


\section*{Supplementary material}

Supplementary Information for

# Estimating Parameter Fields in Multi-Physics PDEs from Scarce Measurements

Xuyang Li, Mahdi Masmoudi, Rami Gharbi, Nizar Lajnef[*], and Vishnu Naresh Boddeti[*]

Michigan State University, East Lansing, MI 48824, USA

*To whom correspondence should be addressed; Email: lajnefni@msu.edu and vishnu@msu.edu.

## Table of Contents

**Supplementary Note 1: Details of Problem Setup**

**Thermal runaway in batteries.** The heat conduction equation governs heat exchanges occurring within the battery. The energy balance equation is directly proportional to the battery concentration rate. Finally, the chemical reaction equation, following the Arrhenius law, regulates both the chemical concentration of lithium in electrolytes and the reaction itself. Furthermore, it is important to note that equation does not solely account for the entirety of generated heat. We specifically address the primary contributing factor of electrolyte decomposition as the predominant source of heat generation responsible for rapid temperature increases during the thermal runaway (TR) process. Variable $\dot{Q}_{exo}$ represents the energy release, $\rho = 2,231.2 kg/m^3$ signifies the battery density, $H = 620 kJ/kg$ is the specific heat release, $W = 335 kg/m^3$ is the specific active material content, $E = 241 kJ/mol$ represents the activation energy, $R_c = 8.314 J/(mol \cdot K)$ denotes the gas constant, and concentration $c$ is a state variable.

Both prismatic cylinder batteries are considered isotropic in the in-plane directions (cross-section A-A) of the battery, wherein $k_y$ and $k_z$ denote identical thermal conductivity. The cross-plane (cross-section B-B, the radial direction in cylinder battery) thermal conductivity is noted as $k_x$, with a relatively lower magnitude compared to the in-plane thermal conductivity. The boundary condition and initial conditions are defined in the equations below.

$$k\nabla T = h_c(T_{amb} - T) \tag{1}$$

$$T(x, y, z, t = 0) = T_0 \tag{2}$$

$$c(x, y, z, t = 0) = c_0 \tag{3}$$

where the heat convection $h_c$ is constant between the battery and the external environment. $T_{amb} = 393.15\ K$ and $h_c = 12\ W/(m^2 \cdot K)$ are the constant ambient environment temperature and the convection coefficient of the heat transfer, respectively. The initial temperature is at room temperature $T_0 = 298.15\ K$ ($25°C$) and the concentration of the lithium is $c_0 = 1$.

To generate reference parameter fields, exponential, quadratic, and cubic relationships with temperature are assumed, building upon previous findings [1,2,3]. Three distinct neural network branches are constructed to model $C_p$, $k_z$, and $k_{xy}$, respectively, taking temperature values from the corresponding spatial domain as input. Neural networks are then trained to minimize errors between measured and predicted temperatures at the surface, with all three field parameters jointly estimated.

Notably, an additional weight is manually added to represent scalar parameter $A_e$ within the neural network. This scalar parameter $A_e$ is jointly estimated using a single neuron with a larger learning rate. When jointly estimating for the cylinder battery, additional observations are provided for the next 40min to exclusively facilitate continuous training of $A_e$. This enables accurate estimation of parameter $A_e$ because it directly influences chemical reactions. Chemical reactions start solely from this time region onwards.

**Flow in porous media.** The general flow problem encompasses diverse processes such as subsurface fluid flow or groundwater flow through soil or aquifers, as well as carbon sequestration and storage. In this study, the flow problem exemplifies the complexities of characterizing porous media and understanding transport phenomena within them. The advection-dispersion equation (ADE) describes the concentration of particles based on the flow velocity $\boldsymbol{v}$, and Darcy flow characterizes the fluid movement through porous media and establishes the relationship between hydraulic conductivity $K$ and hydraulic head $h$. In ADE, $\boldsymbol{v}$ is the average pore velocity in the $x$ and $y$ direction in the 2D space, porosity $\phi$ is 0.317. Dispersion coefficient $\boldsymbol{D}$ is given as $\boldsymbol{D} = D_w \tau \boldsymbol{I} + \boldsymbol{\alpha} \|\boldsymbol{v}\|_2^2$, with diffusion coefficient $D_w = 0.09 m^2/hr$, the tortuosity of the medium $\tau = 0.681$. Dispersivity $\boldsymbol{\alpha}$ is a diagonal matrix with the principal components $\alpha_L = 0.01m$, and $\alpha_T = 0.001m$.

The initial condition for ADE is defined as:

$$u = \begin{cases} \exp(-1600(x - L_x/2)^2), & \text{if } y = 0 \\ 0, & \text{otherwise} \end{cases} \tag{4}$$

and Neumann boundary conditions of $u$ are applied to all 4 boundaries. For the steady state Darcy flow, the boundary conditions are:

$$h(x, y = 0, t) = 0 \tag{5}$$

$$-K\partial h(x, y = L_y, t)/\partial t = 1 \tag{6}$$

$$-K\partial h(x = 0 \text{ or } L_x, y, t)/\partial t = 0 \tag{7}$$

Using a reference field of $K$, the corresponding PDEs are solved numerically from 0 to 16 minutes for a total of 50 timesteps, each timestep takes approximately 20 seconds. We utilize a single feed-forward neural network branch to estimate $K(x,y)$, taking 2D spatial coordinates as input due to its spatial dependency. The neural network model is trained to minimize the error between the predicted concentration and the ground truth.

**Cardiac electrophysiology.** In the Aliev-Panfilov model, $a$ and $b$ are known scalars related to the tissue excitation threshold and refractoriness. The state variable, transmembrane potential $V$, represents the membrane voltage and is accessible in experimental measurements. $W$ denotes an unknown state variable. $k_0 = 8, \mu_1 = 0.2, \mu_2 = 0.3, a = 0.01, b = 0.15$, and $\epsilon = 0.002$.

A Neumann boundary condition of zero is applied to all four boundaries. The initial condition for voltage $V$ is set to zero across the entire domain, while the initial condition $W$ is uniformly set to 0.01. The voltage responses $V$ are computed numerically over the first $40 TU$ with a timestep of $1 TU$. Data from the state $W$ are not used in neural network model training. In this problem, a single feed-forward neural network is constructed to estimate this field parameter $D$.

**Adding Gaussian noise to numerical observations.** In all numerical experiments, Gaussian noise is applied to the available measurements to mimic the effect of real-world inaccuracies. Let the $u(t)$ represent the measured state variable, and noise, denoted by $\eta$, is added proportional to its original value.

$$\eta(t) = u(t) * N(0,1) * \zeta \tag{8}$$

where $\zeta$ is the noise level (i.e. 0%, 1%, 10%, etc.). Thus, the noisy source $\tilde{u}(t)$ can be denoted as:

$$\tilde{u}(t) = u(t) * (1 + N(0,1) * \zeta) \tag{9}$$

Notably, for battery modeling, the temperature modification is performed using the Celsius scale.

**Supplementary Note 2: Additional Results**

**Supplementary Note 2.1: Battery Thermal Runaway**

**Predictions under various initial and boundary conditions.** In Fig. 3 of the article, accurate estimation of battery parameters is demonstrated, and the thermal and chemical behaviors of the batteries exhibit very low errors in prediction. It is important to note that in the main article, we exclusively present the parameter estimation results for the prismatic battery. However, in *Supplementary Figure 1a*, we illustrate the parameter estimation results for the cylindrical battery, where $k(r)$ denotes the thermal conductivity in the radial direction.

Following parameter estimation, a thorough examination of the state variable responses, including temperature and concentration, is conducted under the influence of changes in conditions. The initial temperature decreases from $298.15K$ to $273.15K$, while the ambient temperature (boundary condition) is set 10 degrees higher at $433.15K$ compared to the training condition. In *Supplementary Figure 1b-c*, both the average temperature profile and detailed temperature distribution for the prismatic battery are displayed. No visual discrepancies are observed, and the errors are less than 0.2 degrees. In *Supplementary Figure 1d-e*, analogous outcomes are shown for the cylinder battery. In both instances, our method exhibits high accuracy in modeling the forward response, with minimal observed error.

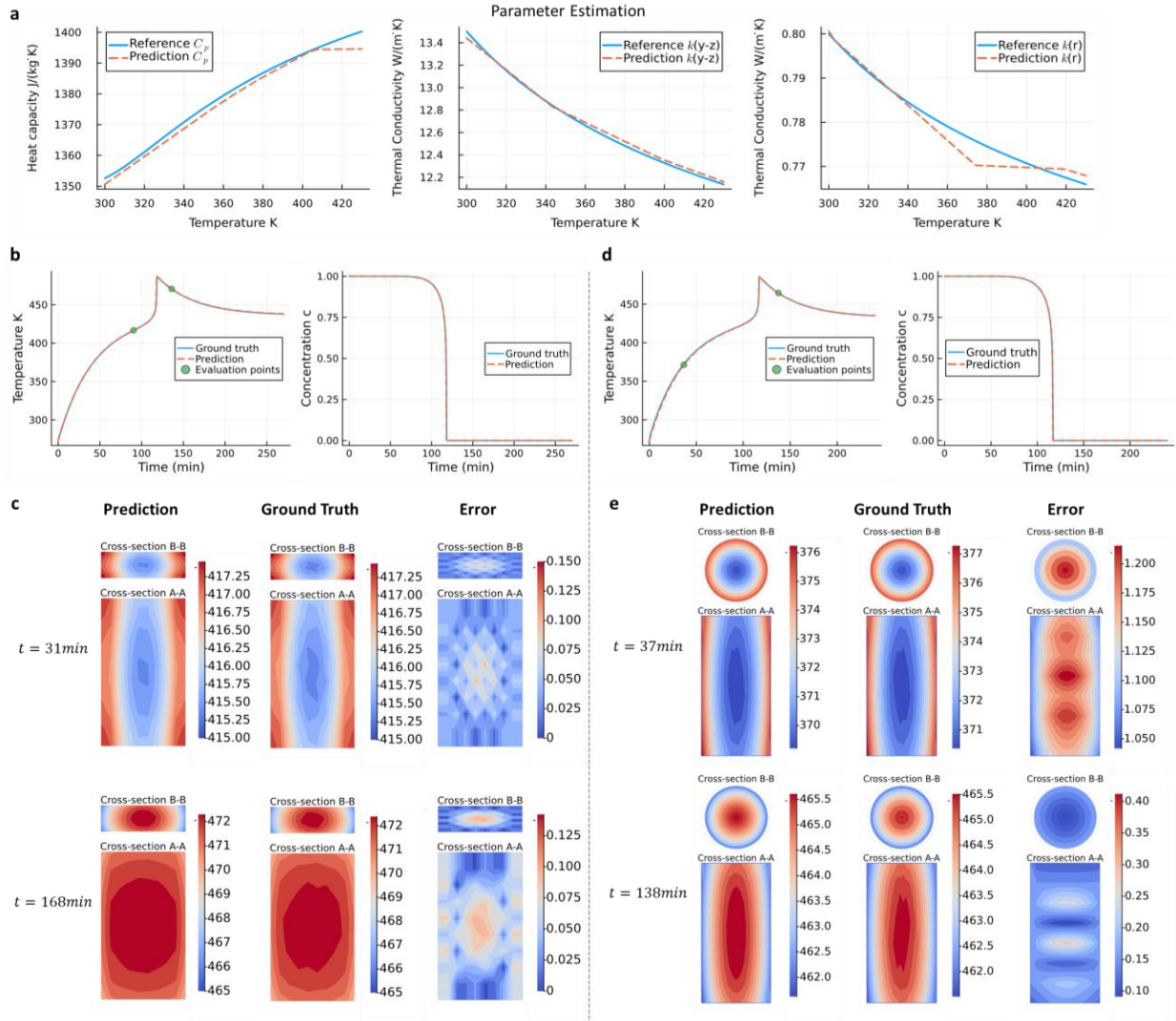

*Supplementary Figure 1.* Thermal runaway predictions under different conditions. **a**. Parameter estimation results for the cylinder battery. **b**. The thermal runaway behaviors with time under different initial and boundary conditions for the prismatic battery. **c**. The temperature profiles for the prismatic battery. **d**. The thermal runaway behaviors for the cylinder battery. **e**. The temperature profiles for the cylinder battery.

**Parameter estimation under varying training samples.** Our proposed method demonstrates robust parameter estimation performance across a range from very scarce samples to hundreds of samples. Using the prismatic battery configuration as a case study, we conduct extensive estimations with varying neural network initializations. *Supplementary Figure 2* illustrates the corresponding parameter estimation errors for heat capacity and thermal conductivity.

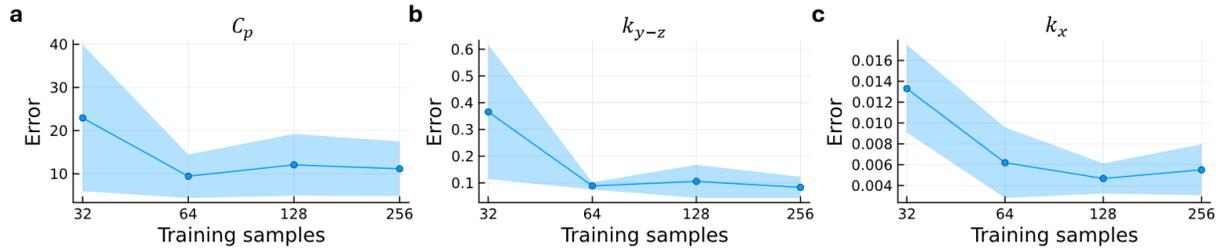

*Supplementary Figure 2.* Parameter estimation performance under different training samples. **a**. Heat capacity. **b**. Thermal conductivity in $x - y$ direction. **c**. Thermal conductivity in $z$ direction. The error band represents the standard deviation from multiple training scenarios.

**Parameter estimation results compared to PINN.** *Supplementary Figure 3* provides a focused evaluation of the PINN framework applied to a representative battery thermal runaway problem. The implementation of PINNs is computationally expensive and demands a large amount of data. Due to the high dimensionality of the original problem, we present a comparative analysis using a simplified version of the battery thermal runaway scenario. Specifically, we reduce the geometry from 3D to a 2D setup. Despite this simplification, training still requires a significant amount of time on an NVIDIA RTX A6000 GPU, approximately 22hours.

While the PINN demonstrates reasonable performance in estimating the forward solution, it struggles with parameter inference and fails to generalize under extrapolative conditions. This highlights the limitations of PINNs in parameter estimation and extrapolation, even in a reduced-dimensional geometry, and underscores the superiority of our method in these aspects.

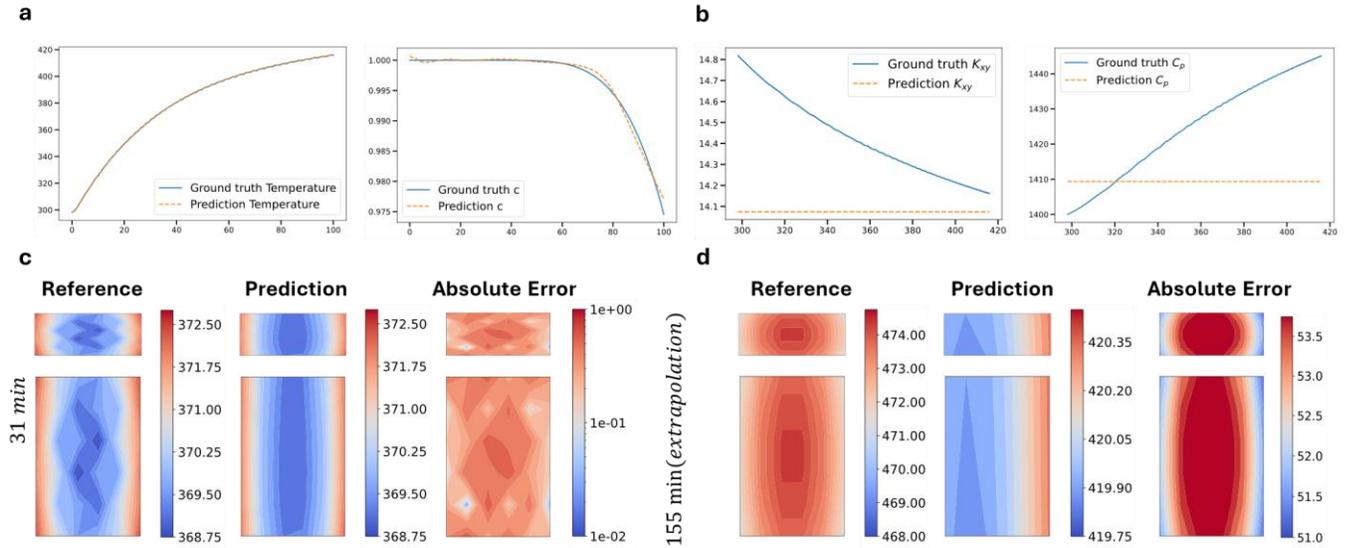

*Supplementary Figure 3.* Parameter estimation results and forward response estimation using PINN for the battery problem.

## Supplementary Note 2.2: Cardiac Electrophysiology

**Parameter estimation results compared to PINN.** *Supplementary Figure 4* presents additional parameter estimation results under extremely scarce training data, using the parameter distribution from Case 1 in the article. As shown, while PINN exhibits a similar error magnitude (in terms of MAE), its overall distributions lack useful alignment with the reference parameters. In contrast, our proposed method demonstrates robust and accurate estimations across all scenarios.

## Supplementary Note 2.3: Flow in Porous Media

*Supplementary Figure 5* presents the results of the parameter estimation task using the Physics-Informed Neural Network (PINN) method. These results support findings previously reported in the literature, showing that PINNs can effectively reproduce the overall physical response with a relatively low level of error. However, the method demonstrates significant limitations in accurately estimating parameters when they are defined as spatially varying fields. Comparative results from our proposed approach are provided in the main paper.

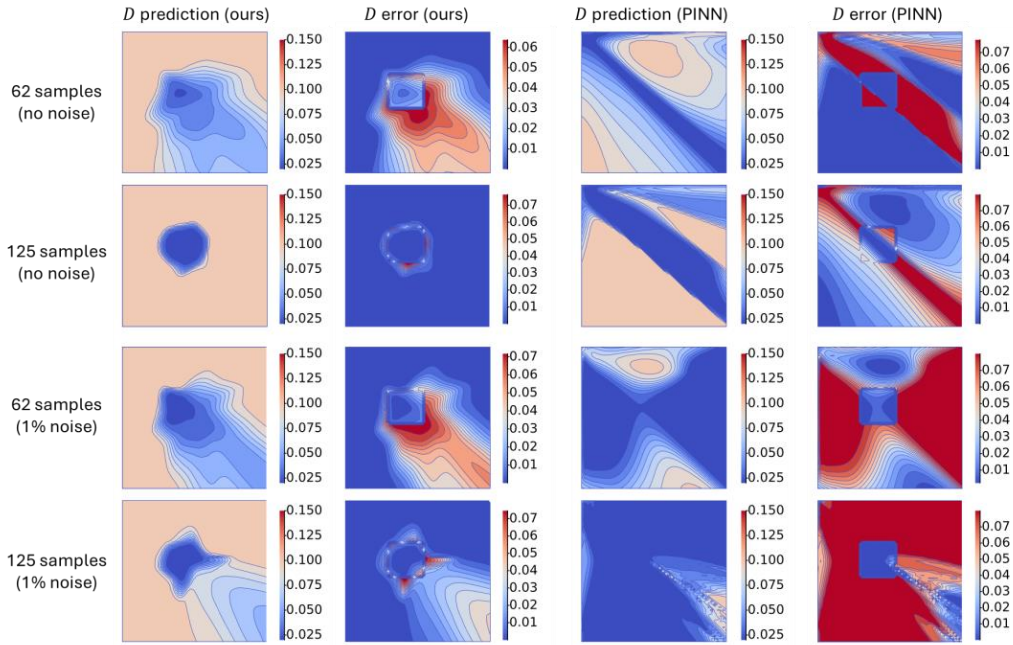

*Supplementary Figure 4.* Parameter estimation results under different training samples and noise levels. The results are compared to PINN.

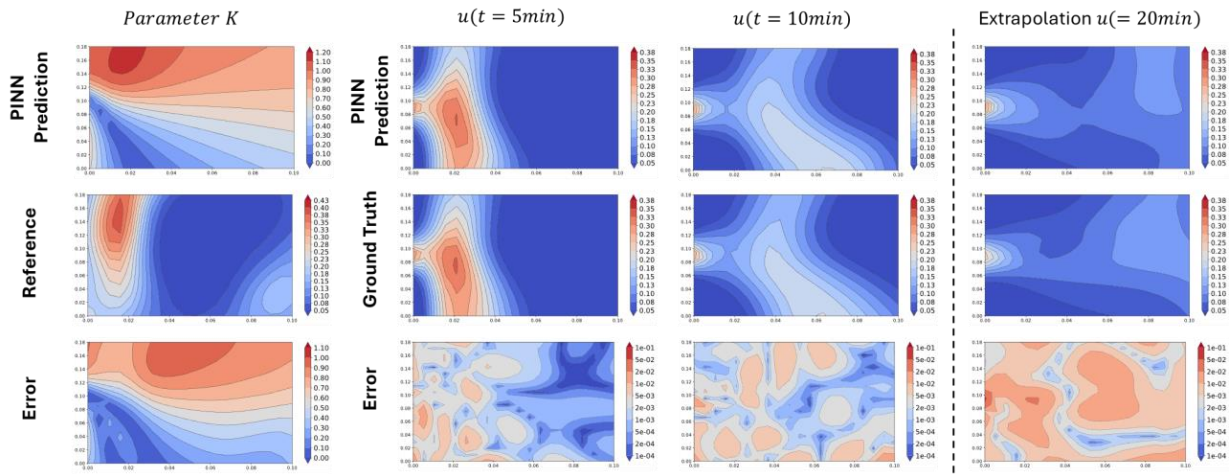

*Supplementary Figure 5.* Parameter estimation results of the PINN method.

## Supplementary Note 2.4: Cell Migration and Proliferation

In the article, we present parameter estimation results for only one set of experiments (cases 1). In the

following figures (*Supplementary Figures 6-8*), we display the results for the other three experimental

datasets (cases 2-4). Notably, in case 3, our parameter estimations, both field and scalar, do not always align with or may show significant discrepancies compared to the results from PINN-SR. However, the predicted density distribution is considerably more accurate, demonstrating that our method can identify a better combination of parameters, leading to more precise modeling of the physical phenomena. We summarize the improvement in density prediction (measured in median errors compared to PINN-SR) in the following *Supplementary Table 1*, considering both scalar parameter estimations and field parameter estimations.

*Supplementary Table 1.* Improvement in modeling cell migration and proliferation phenomena (compared to PINN-SR).

|  | Case 1 | Case 2 | Case 3 | Case 4 |
|---|---|---|---|---|
| Scalar parameters | 20.2% | 28.9% | 15.0% | 18.9% |
| Filed parameters | 44.0% | 47.0% | 36.3% | 27.8% |

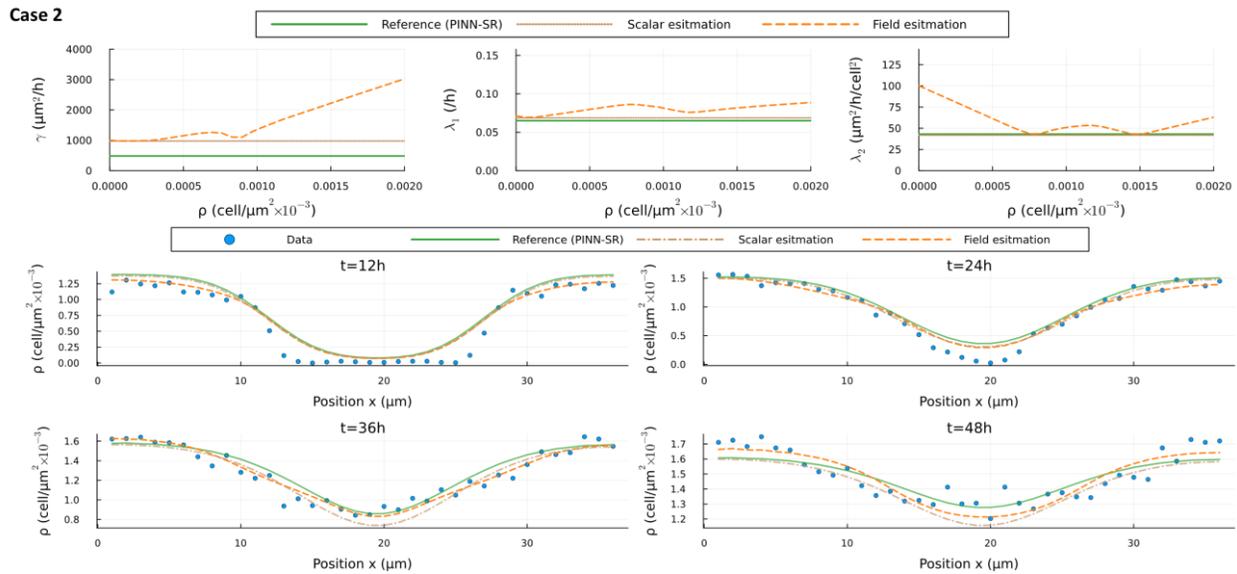

*Supplementary Figure 6.* Parameter estimation results for case 2 in cell migration and proliferation problem.

**Case 3**

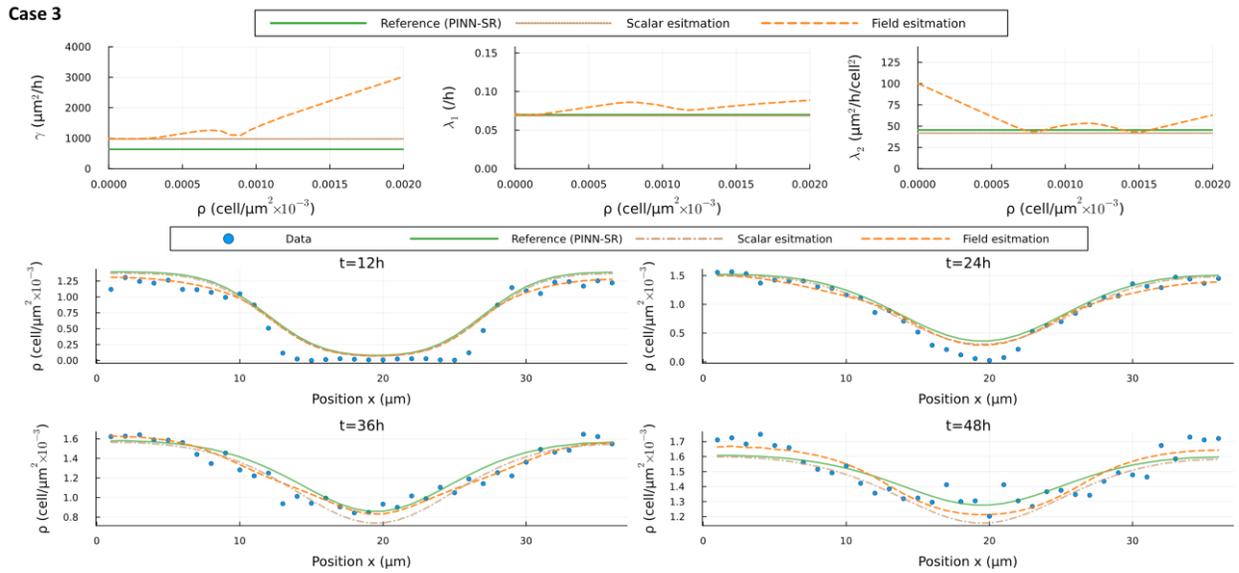

*Supplementary Figure 7.* Parameter estimation results for case 3 in cell migration and proliferation problem.

**Case 4**

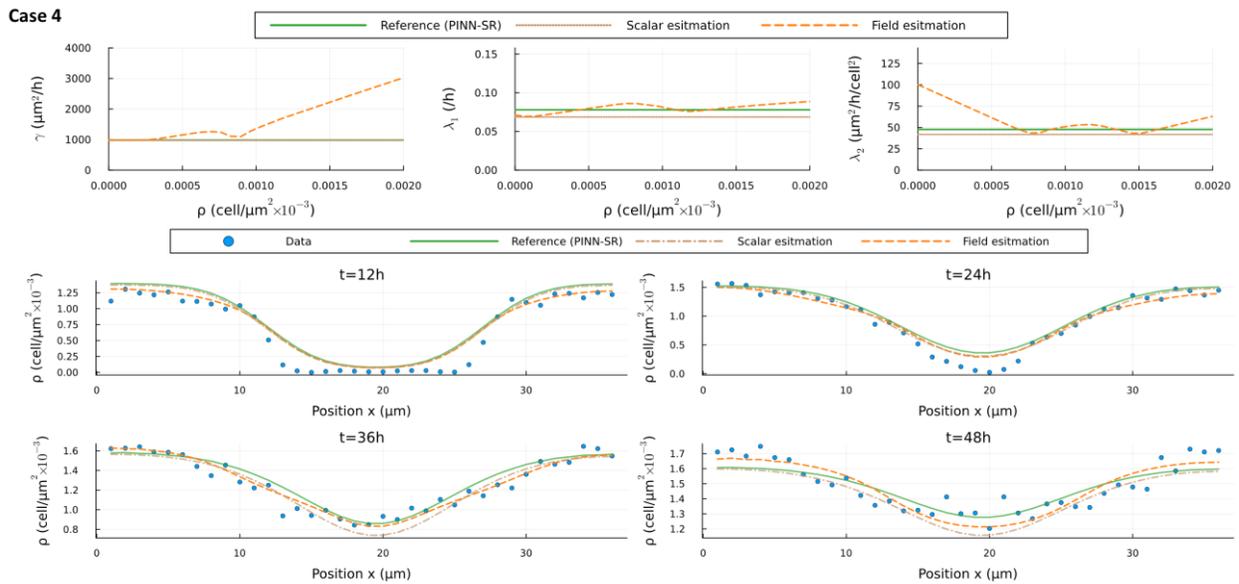

*Supplementary Figure 8.* Parameter estimation results for case 4 in cell migration and proliferation problem.

## Supplementary Note 3: Finite Difference Method for Space Discretization

**Method of lines.** Method of lines (MOL) is a numerical technique for solving PDEs. The first step of MOL is spatial discretization using the finite difference method. Considering the heat conduction problem, the

temperature profile $T$ in one dimension is discretized into $T_1, T_2, \cdots, T_n$. Thus, the second-order derivative of $T$ can be written numerically as:

$$\nabla(\nabla T) = \frac{\partial^2 T}{\partial x^2} \approx \frac{1}{\Delta x^2} \boldsymbol{A}_2^x * \boldsymbol{T} + \boldsymbol{b} \tag{10}$$

$$\boldsymbol{A}_2^x = \begin{bmatrix} -2 & 1 & & & & \\ 1 & -2 & 1 & & & \\ & \ddots & \ddots & \ddots & & \\ & & \ddots & \ddots & \ddots & \\ & & & 1 & -2 & 1 \\ & & & & 1 & -2 \end{bmatrix}; \ \boldsymbol{T} = \begin{bmatrix} T_2 \\ T_3 \\ \vdots \\ T_{n-2} \\ T_{n-1} \end{bmatrix}; \ \boldsymbol{b} = \begin{bmatrix} T_1/\Delta x^2 \\ 0 \\ \vdots \\ 0 \\ T_n/\Delta x^2 \end{bmatrix} \tag{11}$$

where $\frac{1}{\Delta x^2} \boldsymbol{A}_2^x * \boldsymbol{T} + \boldsymbol{b}$ substitutes the second-order spatial derivates, $\boldsymbol{A}_2^x$ is an $n \times n$ banded matrix, $\boldsymbol{b}$ represents boundary value terms, where elements in its first and last rows stem from boundary condition. $\Delta x$ is the grid spacing for the discretization. It is noteworthy that only $n-2$ points, excluding the two boundary terms, are encompassed in equation (1). This is because the boundary terms $T_1$ and $T_n$ in the vector $\boldsymbol{b}$ can be directly derived using the given boundary conditions.

In addition to numerically substituting the second-order derivative, first-order derivatives ($\nabla T$) and higher-order derivatives in other types of problems can be tackled similarly. After the space discretization process, the PDE yields a set of time-dependent ODEs, which can be solved using differential equation solvers.

**Boundary conditions.** This work mainly encounters Dirichlet, Neumann, and a mixed boundary condition known as the heat flux boundary condition. In the following, all three types of boundary conditions are analyzed in the thermal problem to solve for the boundary terms $T_1$ and $T_n$.

- In the Dirichlet boundary condition, the boundary values are explicitly defined for the left side ($\alpha_L$) and right side ($\alpha_R$).

$$T_1 = \alpha_L; \ T_n = \alpha_R \tag{12}$$

- In the Neumann boundary condition, the boundary values are expressed as spatial derivatives.

$$\frac{\partial T_1}{\partial x} = \alpha_L; \frac{\partial T_n}{\partial x} = \alpha_R \tag{13}$$

In this case, forward and backward differences are utilized to derive equation (4) further.

$$\frac{-T_1 + T_2}{\Delta x} = \alpha_L; \frac{-T_{n-1} + T_n}{\Delta x} = \alpha_R \tag{14}$$

$$T_1 = T_2 - \alpha_L \Delta x; T_n = \alpha_R \Delta x + T_{n-1} \tag{15}$$

The term $\frac{1}{\Delta x^2} \boldsymbol{A}_2^x * \boldsymbol{T} + \boldsymbol{b}$ from equation (1) is re-organized by plugging in $T_1$ and $T_n$ values from equation (6), and the corresponding matrix $\boldsymbol{A}_2^x$ and vector $\boldsymbol{b}$ are modified as:

$$\boldsymbol{A}_2^x = \begin{bmatrix} -1 & 1 & & & & \\ 1 & -2 & 1 & & & \\ & \ddots & \ddots & \ddots & & \\ & & \ddots & \ddots & \ddots & \\ & & & 1 & -2 & 1 \\ & & & & 1 & -1 \end{bmatrix}; \boldsymbol{b} = \begin{bmatrix} -\alpha_L/\Delta x \\ 0 \\ \vdots \\ 0 \\ \alpha_R/\Delta x \end{bmatrix} \tag{16}$$

- In the third boundary condition, the expression with the first-order accuracy is:

$$-k_1 \nabla T = h_c(\alpha_L - T_1); k_n \nabla T = h_c(\alpha_R - T_n) \tag{17}$$

$$\frac{-k_1(-T_1 + T_2)}{\Delta x} = h_c(\alpha_L - T_1); \frac{k_n(-T_{n-1} + T_n)}{\Delta x} = h_c(\alpha_R - T_n) \tag{18}$$

where $k_1$ and $k_n$ are the thermal conductivity at the first and last points. $h_c$ is the constant convection coefficient. $\alpha_L$ and $\alpha_R$ are the ambient environment temperatures at the left and right sides of the boundary. The above equation (9) can be further used to derive $T_1$ and $T_n$.

$$T_1 = \frac{\beta_L T_2 + \alpha_L}{\beta_L + 1}; T_n = \frac{\beta_R T_{n-1} + \alpha_R}{\beta_R + 1} \tag{19}$$

where:

$$\beta_L = \frac{k_1}{\Delta x \cdot h_c}; \beta_R = \frac{k_n}{\Delta x \cdot h_c} \tag{20}$$

Consequently, terms $T_1$ and $T_n$ in equation (1) are substituted, and matrix $\boldsymbol{A}_2^x$ and vector $\boldsymbol{b}$ are modified.

$$A_2^x = \begin{bmatrix} \frac{\beta_L}{\beta_L+1}-2 & 1 & & & & \\ 1 & -2 & 1 & & & \\ & & \ddots & \ddots & \ddots & \\ & & & \ddots & \ddots & \ddots \\ & & & 1 & -2 & 1 \\ & & & & 1 & \frac{\beta_R}{\beta_R+1}-2 \end{bmatrix}; b = \begin{bmatrix} \frac{\alpha_L}{(\beta_L+1)\Delta x^2} \\ 0 \\ \vdots \\ 0 \\ \frac{\alpha_R}{(\beta_R+1)\Delta x^2} \end{bmatrix} \tag{21}$$

**Extending to 2D and 3D discretization.** The above equations can be extended to 2D or even 3D cases. In the context of a 2D space, let $T_{ij}$ denote the temperature profile at index $i$ and $j$, where $i = 1,2,\cdots,n$ and $j = 1,2,\cdots,m$. The temperature in the 2D can be vectorized with the following.

$$S_j = \begin{bmatrix} T_{1j} \\ T_{2j} \\ \vdots \\ T_{nj} \end{bmatrix}; T = \begin{bmatrix} S_1 \\ S_2 \\ \vdots \\ S_m \end{bmatrix} \tag{22}$$

where $S_j$ represents the temperature in each column of the 2D temperature profile $T_{ij}$ with a size of $n \times 1$ and $T$ has a size of $nm \times 1$.

Next, the spatial discretization in 2D can be expressed as:

$$\nabla(\nabla T) = \frac{\partial^2 T}{\partial x^2} + \frac{\partial^2 T}{\partial y^2} \approx A_2^{xy} * T + b \tag{23}$$

Here, $A_2$ represents a square matrix of dimensions $nm \times nm$. It can be interpreted as a collection of $m \times m$ blocks, where each block is $n \times n$ in size, as depicted below:

$$A_2^{xy} = \begin{bmatrix} A_2^x - 2D + B & D & & & \\ D & A_2^x - 2D + B & D & \ddots & \\ & \ddots & \ddots & \ddots & \\ & \ddots & D & A_2^x - 2D + B & D \\ & & & D & A_2^x - 2D + B \end{bmatrix} \tag{24}$$

where each $A_2^x$ has a size of $n \times n$ representing the derivative in the $x$ direction corresponding to each $S_j$. The derivative in the $y$ direction is implemented with matrices $D$ and $B$.

$D$ is a diagonal matrix that accounts for the discretization in the $y$ direction. It is defined as $D = I\Delta x^2/\Delta y^2$, where $I$ is the identity matrix with dimensions $n \times n$. $B$ is an extra term that modifies $D$ based on the boundary conditions in the $y$ direction. In addition, $b$ denotes a vector with dimensions $nm \times 1$, obtained through the derivation of boundary conditions.

Furthermore, the above matrix can be expanded to a general 3D case with given boundary conditions, the corresponding matrix $A_2^{xyz}$ size is $nml \times nml$, where $l$ is the number of discretization points in the third dimension.

$$A_2^{xyz} = \begin{bmatrix} A_2^{xy} - 2C + B & C & & & \\ C & A_2^{xy} - 2C + B & C & \ddots & \\ & \ddots & \ddots & \ddots & \\ & \ddots & C & A_2^{xy} - 2C + B & C \\ & & & C & A_2^{xy} - 2C + B \end{bmatrix} \tag{25}$$

**Non-uniform mesh.** Importantly, in the cell migration and proliferation problem, the spatial domain ranges from 0 to $1900 \mu m$ while the 38 measurements are provided at $25, 75, 125, \ldots, 1825, 1875 \,\mu m$. Thus, including the boundaries, there are 40 mesh points are fixed at $x = 0, 25, 75, \ldots, 1825, 1875, 1900 \,\mu m$. The interval is $50 \,\mu m$ for inner points but $25 \,\mu m$ for the boundaries. Thus, a finite difference with non-uniform mesh is employed.

Using forward and backward difference from equation (5) for Neumann boundary condition, we obtain state variable $u_1 = u_2$ and $u_{39} = u_{40}$. The second derivative approximation, for example, at $x_1$ is $\frac{u_2 - 2u_1 + u_0}{h_1 h_0} = \frac{u_2 - u_1}{h_1 h_0} = \frac{2(u_2 - u_1)}{h^2}$, where $h = h_1 = 2h_0 = 50$. Similar equations can be obtained for the other side of the spatial domain. The second derivative approximation for all other mesh points remains unchanged, as the gaps are the same, and $h_0 = h_1 = 50$. We modify equation (2) to obtain the following.

$$\frac{\partial^2 u}{\partial x^2} \approx \frac{1}{h^2} A_2^x * u + b \tag{26}$$

$$\boldsymbol{A}_2^x = \begin{bmatrix} -2 & 2 & & & & \\ 1 & -2 & 1 & & & \\ & \ddots & \ddots & \ddots & & \\ & & \ddots & \ddots & \ddots & \\ & & & 1 & -2 & 1 \\ & & & & 2 & -2 \end{bmatrix} ; \boldsymbol{u} = \begin{bmatrix} u_1 \\ u_2 \\ \vdots \\ u_{37} \\ u_{38} \end{bmatrix} ; \boldsymbol{b} = \begin{bmatrix} 0 \\ 0 \\ \vdots \\ 0 \\ 0 \end{bmatrix} \tag{27}$$

Overall, FDM provides efficient and accurate forward solutions, significantly improving training speed through backpropagation with the adjoint method. However, its geometric limitations restrict it to rectangular or cylindrical domains. While specialized FDMs [4,5,6] can handle complex cases, they often lack generalizability. Future research could explore more flexible techniques, such as the finite element and finite volume methods, for broader applicability. Additionally, FDM introduces numerical errors that can hinder convergence in complex dynamic systems. The trade-off between computational cost and accuracy must be carefully evaluated when applying finite difference modeling.

## Supplementary Note 4: Ablation Study on Neural Network Sizes

In the article, neural network sizes are pre-selected for training. However, the impact of network size cannot be overlooked. In this supplementary note, we investigate parameter estimation and forward response modeling performance with respect to two aspects of neural network size: the number of neurons per layer and the number of residual blocks. The porous media flow case is used as the test scenario for this investigation. The number of training samples is 128. In *Supplementary Figure 9a*, as the number of neurons per layer increases, the estimation error for parameter $k$ decreases. However, at 200 neurons, the errors increase. Despite this, the forward response estimation errors in the second graph stabilize, indicating the model is already well-trained with this number of training samples. Similarly, in *Supplementary Figure 9b*, the forward response estimation stabilizes when using residual blocks, although the parameter estimation errors fluctuate.

Overall, this ablation study demonstrates that even with larger neural networks, the performance of parameter estimation and forward modeling remains unaffected, highlighting the robustness of the proposed method.

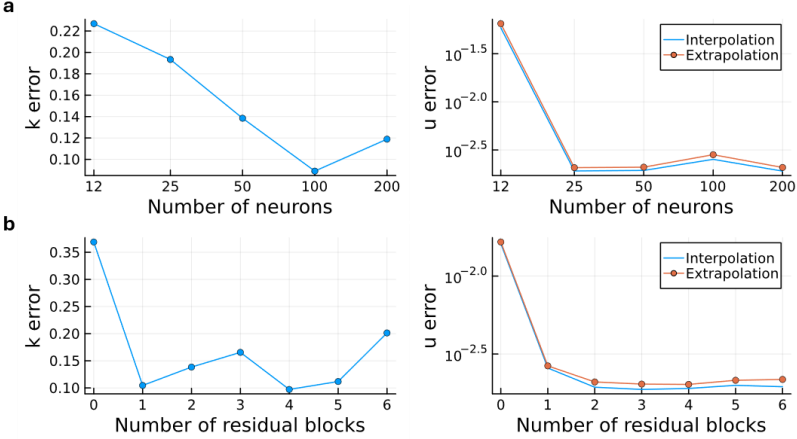

*Supplementary Figure 9*. The effect of neural network sizes on parameter estimation performance.

## Supplementary Note 5: Two-step Training verse other Training Strategies

We conducted comparisons to assess the necessity of the proposed two-step training strategy. The first comparison evaluates using only scalar parameter estimation against the two-step approach. In the cell migration problem, results in both the main article and *Supplementary Note 2.3* confirm that the second-step training for field parameters significantly boosts modeling accuracy (see *Supplementary Table 1*).

The second comparison assesses direct neural network training versus the two-step method. For cell migration, we modeled the physical parameters as $\boldsymbol{\gamma} = \mathcal{N}_1(\rho, \boldsymbol{\theta}_1)$, $\boldsymbol{\lambda}_1 = \mathcal{N}_2(\rho, \boldsymbol{\theta}_2)$, $\boldsymbol{\lambda}_2 = \mathcal{N}_3(\rho, \boldsymbol{\theta}_3)$. Each network's last-layer activation enforces non-negativity. However, even with a small additive term (0.001 or 10% of reference values) to prevent zero predictions, the direct training approach consistently fails to solve the PDEs or failed to train during the adjoint calculation due to improper parameter combinations.

In the third scenario, we consider using a single neural network to predict three parameters together: $[\boldsymbol{\gamma}, \boldsymbol{\lambda}_1, \boldsymbol{\lambda}_2] = \mathcal{N}(\rho, \boldsymbol{\theta}) + [a_1, a_2, a_3]$. Here, the input size is $n \times 1$ and the output size is $n \times 3$. $n$ is the

number of discretization points (also the batch size). Similar to the second comparison, the training process is highly unstable, consistently failing to solve the governing equation.

Overall, these comparisons underscore the importance of sequential parameter optimization. Directly assuming field parameters leads to repeated training failures, while using only scalar parameters falls short of optimal modeling performance.

## Supplementary Note 6: Comparison to Spline-based Parameter Fitting Strategy

In the main article, we employ neural networks (NN) to learn the distribution of field parameters. For comparison, we evaluated alternative fitting techniques, including splines fitting which is a widely used method for data approximation. Splines fitting constructs piecewise polynomial functions that ensure smooth transitions at junctions, maintaining overall smoothness. The key learnable parameters in this method are the knots, which determine the breakpoints where distinct polynomial segments seamlessly connect.

We evaluate the performance of the Spline method for parameter estimation in PDEs, specifically in the flow in porous media problem, using the Dierckx.jl package in Julia. This package provides B-splines (basis splines) and allows us to fit and evaluate 1D and 2D splines efficiently. Our setup consists of field measurements as input and estimated parameter distributions as output, where the spline approximates the spatial distribution of parameters.

We use a fixed number of knots (5,6 for the x and y axes, respectively), chosen based on the trade-off between smoothness and flexibility. A lower number of knots may lead to underfitting, failing to capture key variations, while a higher number of knots increases complexity, making learning unstable and harder to optimize.

**Optimization Approach:** Instead of solving a tridiagonal system explicitly, Dierckx.jl optimizes the spline by minimizing the following objective function:

$$\sum \left(y_i - S(x_i)\right)^2 + \lambda \int S''(x) dx \qquad (28)$$

- The **first term** ensures interpolation accuracy by minimizing the squared error between the data points and the spline.

- The **second term** penalizes large second derivatives, enforcing smoothness and preventing overfitting.

Instead of using standard backpropagation through a neural network, the optimization in Dierckx.jl adjusts the knot positions and spline coefficients using the least squares or smoothing approach. The gradients are computed with respect to the spline parameters. The adjoint method can still be used for PDE-constrained optimization, but the gradients propagate through the spline structure rather than a neural network.

In multiphysics problems, the Spline method struggles to converge when the number of knots exceeds 5–6, highlighting its sensitivity to increased complexity. A fundamental constraint of this approach is its requirement for state values at every spatial point of the grid, whereas our method is more flexible, utilizing randomly sampled points from different time steps without necessitating full grid coverage. Furthermore, the Spline method requires a minimum of 100 spatial data points from a single time step to achieve convergence, whereas our approach can operate effectively with as few as 32 data points, even if they originate from different time steps. Despite using fewer data points, our method achieves significantly lower errors compared to the Spline method, which exhibits larger errors despite its denser sampling requirements. A summary of these observations is provided in *Supplementary Table 2*.

*Supplementary Table 2.* Performance comparison of Spline and Neptune.

|  | **Spline** | **Neptune** |
|---|---|---|
| *Minimum Data Points* | 100 (all in a single time step) | 32 (not necessarily same time step) |
| *Convergence Behavior* | Struggles to converge when using more than 5 knots | Stable |

| *Error Magnitude* | 3× larger than Neptune | Small |
|---|---|---|

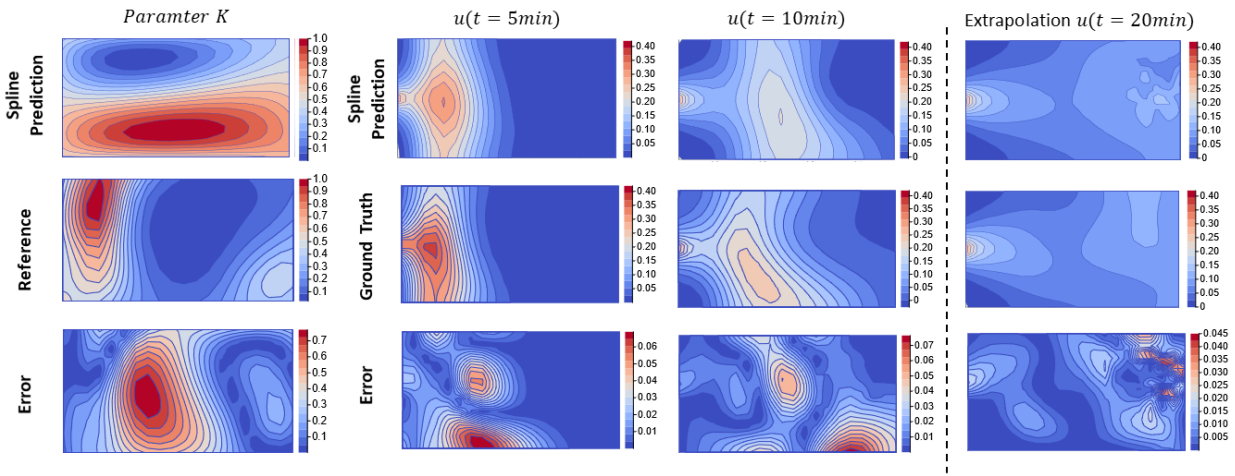

*Supplementary Figure 10.* Spline-based parameter fitting for the flow in porous media problem.

In *Supplementary Figure 10*, we illustrate a representative case for porous media flow. The first column presents the predicted hydraulic conductivity field $K(x, y)$, followed by the particle concentration field $u(x, y)$ at different time steps (5 min, 10min, and extrapolated at 20min). The Spline-based prediction uses 800 data points corresponding to all spatial grid points at two different time steps and shows significant discrepancies compared to the ground truth, particularly in future time steps, as highlighted by the error maps. Our method significantly improves accuracy over this baseline.

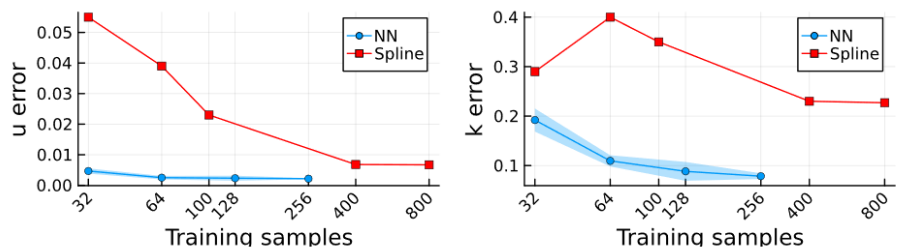

*Supplementary Figure 11.* Comparison of Neural Network and Spline with varying sample sizes.

*Supplementary Figure 11* compares the performance of the Spline-based method with our neural network (NN)-based approach across varying numbers of training samples. Neptune demonstrates significantly better performance, even with as few as 32 randomly distributed data samples across both time and space. In contrast, the Spline method requires at least 400 data points, distributed over a single time step while covering all grid points, to achieve relatively good results. When the Spline method is applied with fewer data points (e.g., 32 or 64), its performance deteriorates substantially. For the case of 100 data points, we sampled every other grid point, leading to a slight improvement, though its performance remained considerably worse than that of our approach. Beyond superior accuracy, our NN-based method also provides greater flexibility in data selection and achieves robust results with significantly fewer samples.

In conclusion, we compared our neural network-based method to the Spline-based approach for parameter fitting in PDE-constrained problems. The Spline method, despite its widespread use in data approximation, struggles with convergence, requires larger spatial sampling, and exhibits larger errors, especially in future steps. Even in the simplest Multiphysics scenario, it fails to match the accuracy and flexibility of our NN-based approach. In contrast, our method achieves significantly lower errors while requiring fewer data points, demonstrating its robustness and scalability for complex parameter estimation tasks.


\clearpage

\end{document}